\documentclass{article}

\usepackage[nonatbib, final]{neurips_data_2024}
\usepackage[dvipsnames,table,xcdraw]{xcolor}
\usepackage{graphicx}
\usepackage[utf8]{inputenc} 
\usepackage[T1]{fontenc}    
\usepackage{hyperref}       
\usepackage{url}            
\usepackage{booktabs}       
\usepackage{amsfonts}       
\usepackage{nicefrac}  
\usepackage{microtype}      
\usepackage{xcolor}         
\usepackage{enumitem}    
\usepackage{soul}
\usepackage{amsmath}
\usepackage{multirow}
\usepackage{caption}
\usepackage{array}
\usepackage{makecell}
\usepackage{subcaption}
\usepackage{wrapfig}
\usepackage{fdsymbol}
\usepackage{booktabs}

\author{
\\[-3em]
\bf Zhecan Wang$^\spadesuit$  \thanks{Equal contribution. Correspondence to: zw2627@columbia.edu} \quad Junzhang Liu$^\spadesuit$ \footnotemark[1] \quad Chia-Wei Tang$^\dagger$ \quad Hani Alomari$^\dagger$ \\[4pt]
\bf Anushka Sivakumar$^\dagger$ \quad Rui Sun$^\spadesuit$ \quad Wenhao Li$^\spadesuit$ \quad Md.~Atabuzzaman$^\dagger$ \quad Hammad Ayyubi$^\spadesuit$ \\[4pt] 
\bf Haoxuan You$^\spadesuit$ \quad Alvi Ishmam$^\dagger$ \quad \quad Kai-Wei Chang$^\vardiamondsuit$ \quad Shih-Fu Chang$^\spadesuit$  \quad Chris Thomas$^\dagger$ \\[4pt]
$^{\spadesuit}$Columbia University \quad
$^{\vardiamondsuit}$UCLA \quad
$^{\dagger}$Virginia Tech\\
\textbf{\url{https://journeybench.github.io/}}
}

\title{JourneyBench: A Challenging One-Stop Vision-Language Understanding Benchmark of Generated Images}

\begin{document}

\maketitle

\begin{figure}[ht!]
    \centering
    \vspace{-2.5em}
    \captionsetup{skip=5pt}    
    \includegraphics[width=\linewidth]{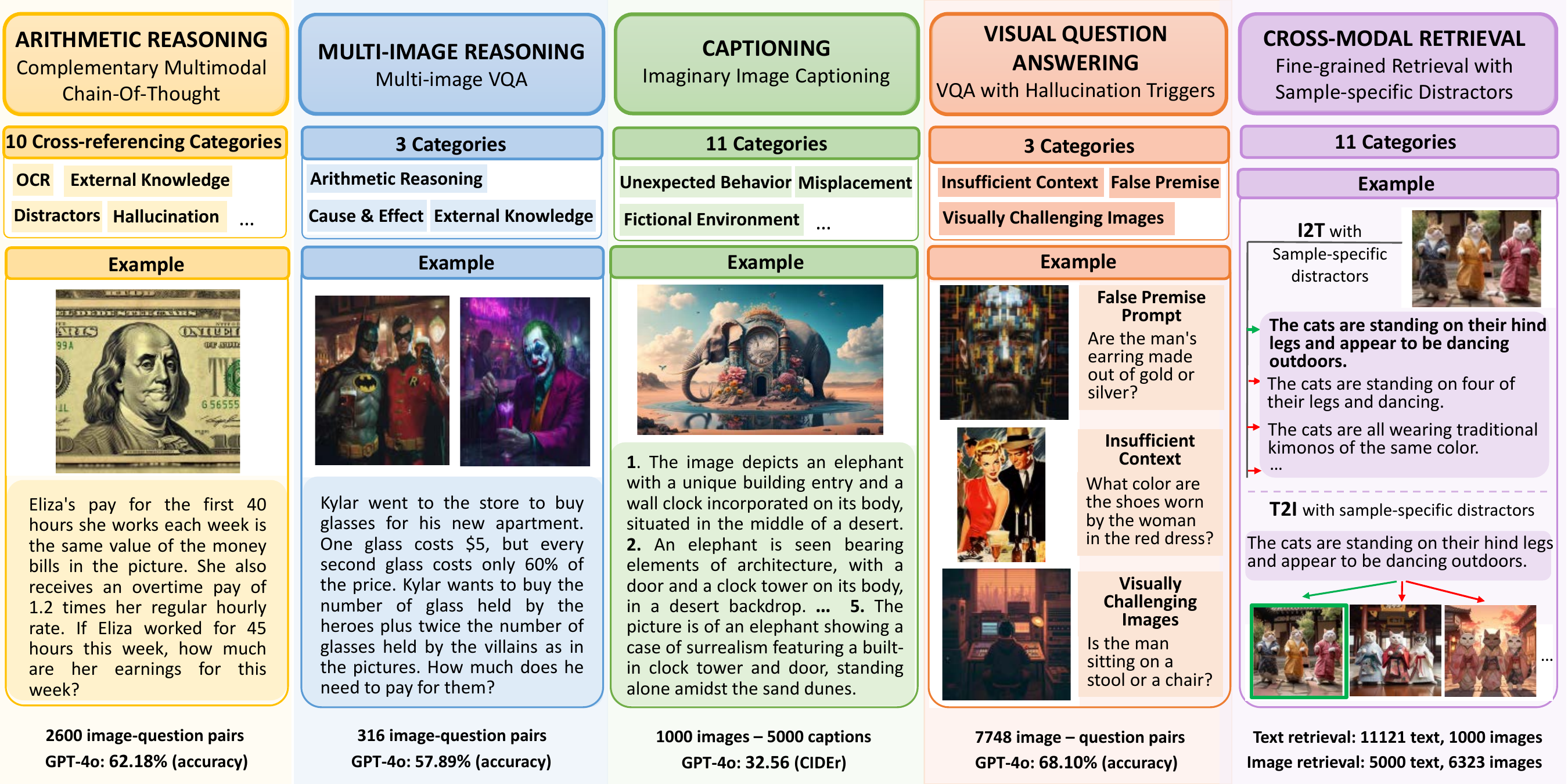}
  \caption{\textbf{JourneyBench Tasks with Fine-grained Categories and Example Data}. JourneyBench includes five fundamental vision-language understanding tasks with unconventional imaginary images to test the limits of models' biases, hallucination tendencies, and fine-grained perception abilities. 
  }
  \vspace{-1pt}
  \label{fig:overview}
\end{figure}

\begin{abstract}
Existing vision-language understanding benchmarks largely consist of images of objects in their usual contexts.
As a consequence, recent multimodal large language models can perform well with only a shallow visual understanding by relying on background language biases. 
Thus, strong performance on these benchmarks does not necessarily correlate with strong visual understanding. 
In this paper, we release JourneyBench, a comprehensive human-annotated benchmark of generated images designed to assess the model's fine-grained multimodal reasoning abilities across five tasks: complementary multimodal chain of thought, multi-image VQA, imaginary image captioning, VQA with hallucination triggers, and fine-grained retrieval with sample-specific distractors.
Unlike existing benchmarks, JourneyBench explicitly requires fine-grained multimodal reasoning in unusual imaginary scenarios where language bias and holistic image gist are insufficient.
We benchmark state-of-the-art models on JourneyBench and analyze performance along a number of fine-grained dimensions.
Results across all five tasks show that JourneyBench is exceptionally challenging for even the best models, indicating that models' visual reasoning abilities are not as strong as they first appear. 
We discuss the implications of our findings and propose avenues for further research.
\end{abstract}

\section{Introduction}

Multimodal large language models (MLLMs) combine the reasoning capabilities of LLMs with visual (and/or other) modalities, enabling them to tackle a wide array of tasks requiring multimodal understanding, such as visual question answering (VQA) \cite{vqav2, vcr, gqa}, multimodal chain-of-thought reasoning \cite{m3cot, zhang2024multimodal}, text-to-image generation \cite{parti, imagen} image captioning \cite{mscoco, flickr}, and so on. 
Their impressive performance has led to rapid adoption in our daily life for various tasks such as mathematical reasoning \cite{mathvista, scienceqa}, navigation \cite{zhou2024navgpt,chen2021history}, and robotic control \cite{driess2023palm,cui2024survey}.
This necessitates their rigorous evaluation before deployment in production systems.

While existing Visual Language Understanding (VLU) benchmarks \cite{yue2023mmmu, fu2024mme, liu2024mmbench} have driven significant progress, they mostly contain limited visual diversity and less complex scenarios than encountered in daily life. For example, many benchmarks restrict their image distribution to resources like COCO \cite{mscoco} or Flickr \cite{flickr} due to copyright constraints on internet-harvested images.
As a result, these benchmarks tend to emphasize commonly occurring subjects, predicates, and objects, over unusual or abstract scenes. 
This enables models to excel by leveraging previously acquired common-world knowledge without necessarily understanding the actual content of the images. 
While this bias might inflate scores on academic benchmarks, it can lead to significant challenges when transitioning to real-world applications \cite{recht2019imagenet}. 
Moreover, benchmarks curated to evaluate Multimodal Chain-of-Thought (MCOT) reasoning such as \cite{scienceqa}, often feature redundant visual content (i.e.~not needed to answer the question), as illustrated in Figure \ref{fig:mcot_comparison}. 
Current MCOT benchmarks also fail to adequately address critical issues like hallucination \cite{hallusionbench} and prediction consistency. 
On retrieval benchmarks, models' performance is saturating near human-level \cite{mscoco, flickr}, making it challenging to distinguish between models. This saturation is partly due to the lack of fine-grained detail in current retrieval benchmarks, which do not sufficiently challenge today's powerful models \cite{winoground}.


The rise of prompt-based generated images presents a unique opportunity for a comprehensive multimodal benchmark. Unlike real images, these generated images bypass copyright issues and offer diverse visual content, enabling more challenging and nuanced testing scenarios. Generated images can combine uncommon concepts, such as ``elephant on macaroons'' which are rare in traditional datasets but critical for evaluating a model's true understanding of visual concepts. 
For example, COCO contains object relations found in ConceptNet \cite{liu2004conceptnet} 68\% of the time vs. only 6\% in the generated images we collect.
Further, as generated images become increasingly realistic and proliferate online, incorporating them into benchmarks for assessing models' capabilities to understand and interpret diverse visual scenes will become increasingly important. 
By leveraging prompt-based generated images, we can address the limitations of existing benchmarks, providing better controllability and diversity in visual content.
This approach enables rigorous testing of models' hallucination tendencies, consistency, and ability to function effectively in varied and unpredictable environments.

With this insight, 
we present \textbf{JourneyBench}, a comprehensive VLU benchmark leveraging prompt-based generated images within a novel human-machine-in-the-loop (HMIL) framework. While some recent works leveraging generated images have been proposed, they are either on a small scale \cite{whoops} (e.g.\textasciitilde1K samples) or not challenging and comprehensive enough  \cite{journeydb}. In contrast, JourneyBench is large (\textasciitilde 13.5K samples) and evaluates models' advanced reasoning capabilities across five challenging tasks: MCOT, multi-image MCOT (MMCOT), fine-grained cross-modal retrieval (CR), open-ended visual question answering (VQA) with hallucination triggers\footnote{
Similar to other recent benchmarks \cite{mathvista}, JourneyBench builds on top of a prior, unpublished benchmark (by the authors) for VQA with hallucination triggers called HaloQuest. We include a complete write-up in our supp.~and do not repeat details here. All other components of JourneyBench are new and described herein.}, and imaginary image captioning. It specifically assesses models' hallucination tendencies, prediction consistency, and ability to understand and differentiate fine-grained details. Our contributions are as follows:

\begin{itemize}[noitemsep, nolistsep, leftmargin=1em]
    \item We introduce JourneyBench, a comprehensive, expertly annotated, challenging VLU benchmark of imaginary images to rigorously test models' capabilities across five tasks.
    \item To the best of our knowledge, for the first time, we address VLU evaluation with  imaginary (unusual or fictional) images on a large scale. 
    We further contribute the challenging complementary MCOT, nvoel multi-image MCOT and fine-grained retrieval tasks with generated images.
    \item We develop a novel adversarial HMIL framework to scale up the generation of high-quality data.
    \item We conduct detailed analyses to provide insights into model performance, behavior and limitations. For instance, even the powerful model GPT-4, achieves only 57.89$\%$ accuracy on multi-image VQA and struggles with co-referencing across modalities in MCOT, achieving just 62.18$\%$ accuracy. 
\end{itemize}





\section{Related Works}



VLU evaluation has been a crucial tool in assessing AI performance across various tasks\cite{zhang2024visionlanguage}, including cross-modal retrieval \cite{mscoco, flickr, eccv-caption}, MCOT \cite{scienceqa, m3cot, mathvista, zhang2024multimodal, yue2023mmmu}, image captioning \cite{mscoco, flickr}, visual question answering (VQA) \cite{vqav2, gqa, vcr, okvqa, aokvqa, visitbench}, and multi-image visual reasoning \cite{multivqg, mantis, mementos}. Despite their significance, there have been limited efforts \cite{whoops, journeydb} to leverage generated images in VLU evaluation. These attempts have not fully exploited the controllability, convenience, and strengths of prompt-based generated images \cite{rombach2022highresolution, midjourney} to address more challenging issues such as MCOT, fine-grained cross-modal retrieval \cite{wang2022paired, tokenflow}, and multi-image visual reasoning \cite{multivqg, mantis, mementos}. Cross-modal retrieval is a fundamental capability of AI with applications in many domains \cite{zhu2023cross}. However, recent models' performances have plateaued on existing benchmarks \cite{mscoco, flickr, eccv-caption}, which primarily focus on differentiating non-related image-text pairs. This allows models to succeed by memorizing holistic styles or content without paying attention to fine-grained visual details \cite{wang2022paired, tokenflow}. Our fine-grained multimodal retrieval task, on the other hand, uses prompt-based generated images to create sample-specific distractors, challenging models to differentiate intricate details. MCOT is another challenging task that involves reasoning across visual and textual modalities. Existing VQA and MCOT datasets often include redundant images, allowing models to solve problems using text inputs alone \cite{wang2023dataset, scienceqa, mathvista}. Furthermore, these datasets fail to address hallucination and consistency issues in real-world math problems \cite{ho2023large, magister2023teaching, Ji_2023, huang2023survey, zhang2024multimodal}. To tackle these limitations, we develop complementary MCOT questions that require the integration of information from both modalities. Additionally, by pairing the same math reasoning question with different visual contexts, we can assess models' consistency and behavior, leveraging the flexibility of generated images. While many existing datasets for image captioning \cite{mscoco, flickr, eccv-caption} and VQA \cite{vqav2, gqa, vcr, okvqa, aokvqa, visitbench} focus on everyday scenarios with real images, our tasks—imaginary image captioning and HaloQuest \cite{haloquest} —aim to evaluate models' understanding of imaginary images, including unusual and fictional visual scenes. By harnessing the strengths of prompt-based generated images, we enhance these popular VLU tasks to push the boundaries of benchmarking high-performing models.


\section{JourneyBench}
In this section, we discuss the procedure for constructing JourneyBench. 
We first describe our approach to collecting high-quality, diverse, and interesting images. Then, we detail the annotation process for each of the five tasks. 
We include further details of our dataset, like quality assurance via multiple rounds of annotations, consistency checks, and dataset statistics, in the appendix. 
Collectively, JourneyBench's curation involved over \textit{2,200 hours} of human annotation effort.


\subsection{Data acquisition and filtering}
\noindent \textbf{\textit{Retrieving generated images.}}
We aim to create a VLU benchmark containing challenging and diverse imaginary images, including unusual, abstract, and complex ones by leveraging the advantages of prompt-based generated images. However, generated images tend to suffer from low quality and biased distribution.
To prevent that, we instead \textit{retrieve} popular prompt-based generated images from Midjourney \cite{midjourney}
- a large crowd-based platform - using web scraping tools with metadata information.
We ensure the diversity of image content by adopting the strategy from \cite{haloquest} -- combining 17 topic words and 15 attribute words to form the query used to retrieve images. This approach results in a significantly larger and more diverse set of topic words for image content compared to previous image-text datasets \footnote{Detailed analysis in Section \ref{sec:qualitative_qnalysis} and appendix.
}.


\noindent \textbf{\textit{Image filtering.}} 
Human annotators select images from the retrieved pool that are: \textbf{unusual}, \textbf{fictional} (unrealistic), and contain visually \textbf{comprehensible} concepts. 
Unusual images depict scenarios outside of everyday experiences, feature unexpected juxtapositions of objects, or include visually striking elements. Fictional images present unrealistic or impossible scenes (\textit{e.g.}, an elephant standing on macaroons). 
Comprehensibility ensures that images are free of artifacts and understandable to humans. 
This balances the fine dynamics between creating challenging scenarios and ensuring legible visual concepts to reliably test models.
We present annotators with a set of questions to help them identify if images fulfill these three criteria.
To address human subjectivity in this task
, we employ at least four Amazon Mechanical Turk (MTurk) annotators for each image. 
They achieve 100\% agreement in over 72\% of cases. 
Detailed information about the user interface, data filtering process, and questions are provided in the appendix.

\noindent \textbf{\textit{Categories of imaginary images}}
Providing a fine-grained categorization of imaginary images can assist in our understanding of models' behaviors across categories of unfamiliar scenarios. Hence, we categorize our images based on how unusual or how unrealistic they are. 
Because of the subjective nature of this problem, we hire four experienced co-author annotators 
who collectively converged on 15 categories of unusualness and unrealisticness across images, as listed in the axes of Figure \ref{fig:cr-captioning-across-categories}, which were then used to annotate the dataset.



We next present how we use imaginary images to form challenging VLU tasks within JourneyBench. 

\begin{figure}[t]
    \centering
    \captionsetup{skip=5pt}
    \scalebox{0.3}{
    \includegraphics{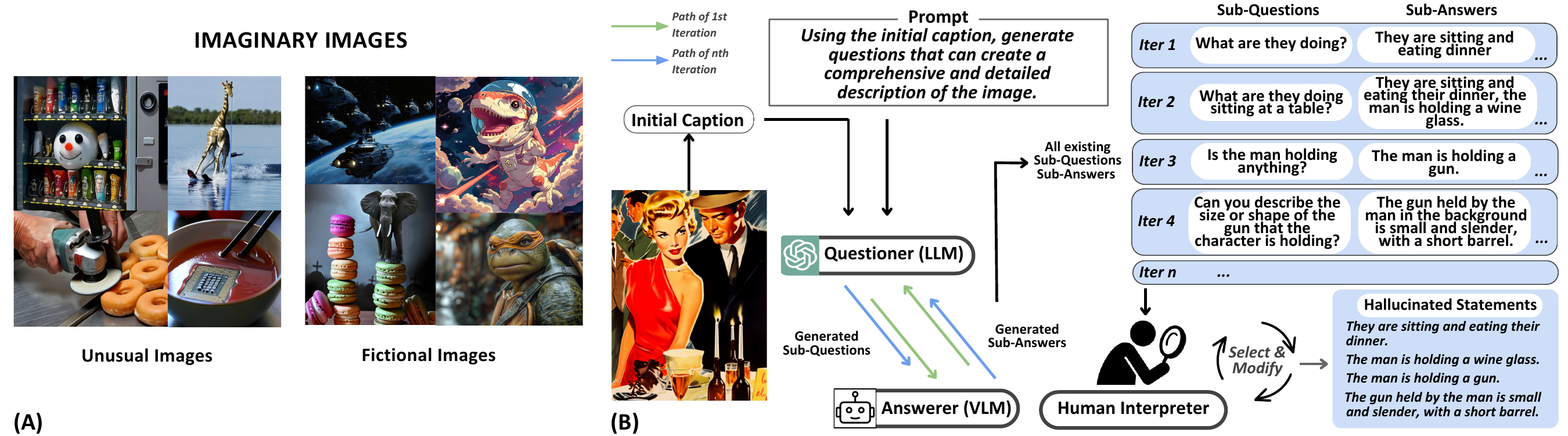}
    }
  \caption{{\textbf{Examples of Imaginary Images} and \textbf{ Human-Machine-in-the-Loop Pipeline.} }}
  \label{fig:hmil_pipeline}
  \vspace{-1.5em}
\end{figure}

\subsection{Imaginary Image Captioning}
\label{sec:captioning}
While captioning is a standard task for VLU benchmarking,
we seek to test models' abilities to understand and caption \textit{imaginary} images in JourneyBench. To this end, we require models to generate a single-sentence description of an image highlighting elements that make it imaginary. 
The ground truth annotation of each collected imaginary image is written by eight MTurk annotators to describe the most unusual or fictional part of the image. Then the captions are verified by another four experienced MTurk annotators to avoid subjective biases among annotators. The user interfaces and detailed procedures during the annotation process are in the appendix. 

\subsection{Fine-grained Cross-modal Retrieval}

Cross-modal retrieval is a fundamental VLU task included in many benchmarks \cite{mscoco,flickr,eccv-caption}. Given an image, the objective is to retrieve the matching text, and vice versa. 
This capability is critical for AI models in various domains, including search engines. However, the performance of existing models on popular cross-modal retrieval benchmarks such as MS-COCO \cite{mscoco} and Flickr30K \cite{flickr} has reached saturation \cite{coco-counterfactual}. These benchmarks primarily involve real images and focus on largely discriminating between pairs holistically. For example, in image-to-text retrieval, other images' matching texts are treated as distractors (i.e.~negatives), even though they are largely irrelevant to the target image, making the task easier.
However, for models to accurately retrieve relevant content, it is crucial to be able to differentiate image-text pairs at a fine-grained level. Thus, to challenge models' ability to perform fine-grained differentiation across similar images, we propose an adversarial HMIL framework to create sample-specific distractors, i.e. hard negatives which require fine-grained discrimination to overcome, for each query sample. For instance, as illustrated in the rightmost examples in Figure \ref{fig:overview}, our framework creates challenging scenarios requiring models to focus on intricate details to successfully retrieve the correct image-text pairs. We next describe our data curation and annotation approach below for each retrieval direction.

\noindent \textbf{\textit{Image-to-Text retrieval.}} 
We experiment with two HMIL approaches to scale up and generate distractors. In the first one, we feed the ground-truth caption (Sec.\ref{sec:captioning}) into MLLMs like GPT-4V and prompt them to generate relevant but conflicting hallucinated statements using in-context examples. Human annotators then verify these generated distractors. This approach is effective but has limitations. It performs well when the image is easily comprehensible by the MLLMs and the ground-truth caption is detailed. 
However, the generated distractors are often not challenging enough and somewhat obvious, as the conflicting elements are ``guessed'' by the generation model, which itself introduces bias. 
We find in cases where the image is complex, or the ground-truth text is not detailed, the model often introduces irrelevant elements into the distractors, reducing their quality.

To address these limitations, we develop a more effective HMIL system inspired by \cite{idealgpt} that introduces a dialogue between an LLM and an MLLM. As in Figure \ref{fig:hmil_pipeline}, the process begins by feeding the initial ground-truth caption and the prompt into the LLM, which generates questions about the image that are answered by the MLLM. 
With each iteration, the MLLM-LLM's errors propagate, making the hallucinated predictions more difficult to overturn and thus revealing ``blind spots'' to humans. These ``blind spots'' are not merely imagined by the generators but empirically demonstrated on the task.
Human annotators then pinpoint these spots, collecting hallucinated answers or statements as potential distractors. 
We found this HMIL approach 
generates high-quality distractors with relevant but conflicting details that are challenging for models to notice.

\noindent \textbf{\textit{Text-to-Image retrieval.}}
Similar to image-to-text retrieval, for each target text, we use the matching ground-truth image to obtain sample-specific image distractors. 
We employ a group of expert annotators to query the Midjourney platform to retrieve relevant but conflicting image distractors for each sample. 
During this process, annotators are asked to find image distractors based on two criteria: the subject, the composition, or both. For example, as illustrated in the bottom rightmost image in Figure \ref{fig:overview}, for the subject criterion, annotators should find image distractors that also feature three cats. For the composition criterion, they should find image distractors where there are three animals positioned side by side and facing the camera. 
By adhering to these criteria, we ensure that annotators collect high-quality image distractors that cannot be easily differentiated without fine-grained details. On average, for each target text, we obtain about five sample-specific distractors.

\subsection{Complementary Multimodal Chain-of-Thought}
In the MCOT task, the input consists of an image and a question which requires the model to integrate information from both modalities. 
However, existing MCOT resources like MathVista \cite{mathvista} and ScienceQA \cite{scienceqa} often contain redundant visual information, allowing models to answer questions using only the language input.
To address this, we aim to build a \textbf{Strictly Complementary} MCOT dataset that \textit{requires} multimodal reasoning. 
In this dataset, visual and text information will be complementary, ensuring models must co-reference both modalities for chain-of-thought reasoning. Our experiments reveal that multimodal co-referencing during the chain-of-thought process is very challenging for existing models. For example, GPT-4 achieves over 90\% accuracy on the text-only version of our COT questions, GSM8K \cite{gsm8k}, but only 49.34\% and 61.2\% in our strictly complementary MCOT setting for GPT-4V and GPT-4o, respectively. This significant drop highlights the importance of our complementary MCOT dataset in evaluating multimodal reasoning capabilities.

\begin{figure}[t]
    \centering
    \captionsetup{skip=5pt}
    \scalebox{0.27}{
    \includegraphics{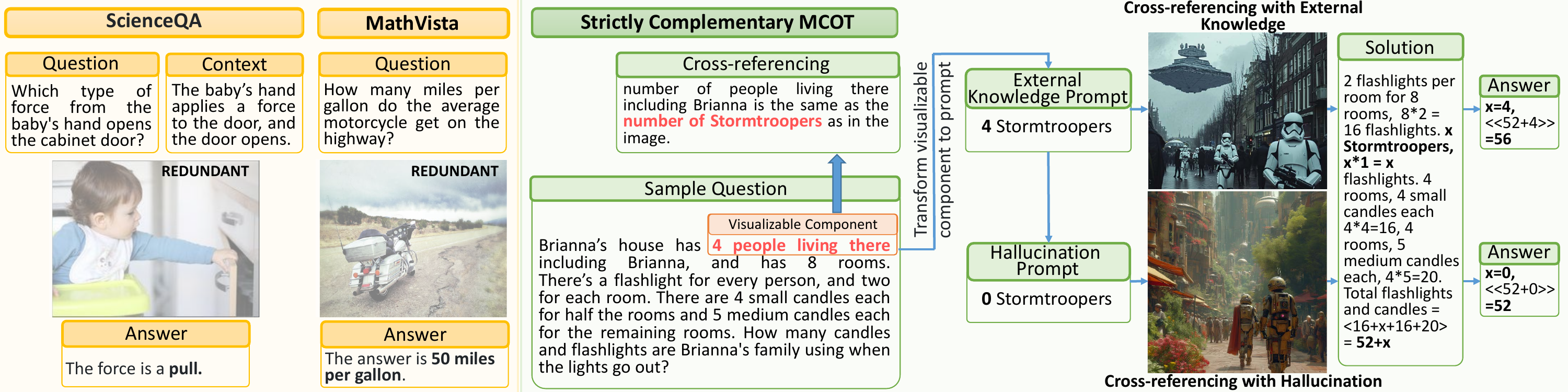}
    }
  \caption{{\textbf{Comparison between ScienceQA, MathVista (left), and our Strictly Complementary MCOT (right) with Examples.} While ScienceQA and MathVista images provide redundant visual information,
  Journeybench provides complimentary visual information that is necessary to answer the question.
  This ensures a more rigorous evaluation of multimodal reasoning capabilities.}}
  \label{fig:mcot_comparison}
\end{figure}
\noindent \textbf{\textit{Visualizing text-only MCOT.}} 
We scale up the generation of strictly complementary MCOT data by converting the text-only COT benchmark, GSM8K \cite{gsm8k}, into MCOT using prompt-based generated images. 
As shown in Figure \ref{fig:mcot_comparison}, the process begins by identifying visualizable text components and converting them into prompts to generate images. 
These images replace the identified text components with new text requiring co-referencing the image. 
This method rapidly scales up the creation of high-quality, complementary MCOT data which allows testing of models' multimodal reasoning capabilities in solving arithmetic problems.


\noindent \textit{\textbf{Co-referencing categories.}} 
Generated images' controllability allows us to test each question with diverse visual contexts all requiring the same arithmetic reasoning logic to better understand models’ abilities. 
As shown in Figure \ref{fig:mcot_across_categories}, we evaluate models' ability to co-reference visual content requiring external knowledge for arithmetic problems and assess hallucination tendencies by omitting referenced objects. 
Despite recent MLLM progress in MCOT benchmarks, co-referencing remains extremely challenging.
We categorize types of co-referencing to analyze models' weaknesses in Figure \ref{fig:mcot_across_categories}. 
Our appendix contains detailed definitions of each type shown. 
Our findings indicate models struggle with hallucination and using external knowledge in the MCOT task, highlighting the need for further research.

\vspace{-4mm}
\subsection{Multi-image Visual Question Answering}
\vspace{-2mm}

Recently, benchmarks for multi-image VQA have been proposed \cite{mantis, tanaka2023slidevqa}, requiring models to reason over multiple images for VQA. However, due to limited real image resources, existing datasets primarily test basic abilities like color matching, image-text matching, and object counting. In contrast, our multi-image VQA task evaluates three specific and challenging reasoning categories: arithmetic reasoning, applying external knowledge to visual reasoning, and identifying cause and effect, as shown in the example of Figure \ref{fig:overview}.

For multi-image VQA data requiring arithmetic reasoning, we use a similar approach to our single-image MCOT data collection. For data requiring external knowledge, we engage six expert annotators to identify and collect high-quality Midjourney images that require external knowledge to understand. 
These annotators then generate multi-image visual questions based on these images. For the cause-and-effect category, we use prompt-based generated images to convert the text-only cause-and-effect dataset, COPA \cite{copa}. Each COPA sample contains two text events representing cause and effect. Annotators identify samples with visualizable events and obtain corresponding generated images, which are then compiled into multi-image samples to test if models can identify the cause or effect between visual events. Our multi-image VQA setting challenges even the best models with complex reasoning tasks requiring co-referencing, applying external knowledge, and understanding cause-and-effect relationships across multiple images.

\section{Experiment}
\subsection{Evaluation Metrics}
For cross-modal retrieval, we report Recall@k (R@k) for $k\in{1,5,10}$. 
For captioning, we report the standard BLEU, ROUGE, CIDEr, and Meteor scores.
For our MCOT and multi-image VQA tasks, 
we use Llama-3-8B \cite{llama-3}
to extract the answers from the models' generated solutions and then again ask Llama-3-8B to determine if the answer is correct by providing the question and ground truth answer with the prompt. We then use Llama-3-70B for solution verification by asking Llama to verify if the generated solution follows the logic of the ground truth solution. We manually verified a subset of Llama-3's responses to ensure quality.
In the appendix, we provide additional details of our evaluation setup, along with the prompts used. 

\subsection{Baseline Models}

For our retrieval tasks, we employ SOTA retrieval pre-trained models, including ALBEF \cite{albef}, CLIP \cite{clip}, $\text{X}^2$-VLM (Large) \cite{x2vlm}, BEiT3 \cite{beit3}, BLIP2 \cite{blip2}, OpenCLIP-Coca \cite{coca}, and InternVL \cite{Internvl}. In the case of MCOT, multi-image VQA, and captioning tasks, we leverage current SOTA vision-language generative models in a zero-shot manner, along with GPT-4o \cite{openai-gpt4-o}
and GPT-4V \cite{openai-gptv}
. The models utilized for these tasks include LLaVA-NeXT \cite{llavanext}, VILA \cite{vila}, BLIP-2 \cite{blip2}, Mantis \cite{mantis}, InternVL \cite{internvlV1.5}, MiniGPT-4 \cite{minigpt}, mPLUG-Owl \cite{mplug}, mPlug-Owl2 \cite{mplugowl2}, Idefics2 \cite{idefics2}, and CogVLM2 \cite{cogvlm}. 
We use different versions and sizes of these models with our fixed prompts, and the details can be found in the appendix.

\begin{table*}[ht]
\centering
\selectfont
\renewcommand{\arraystretch}{1.4}
\resizebox{\textwidth}{!}{
\begin{tabular}{l|cccccccc||cccccccc}
\hline 
\hline
\multicolumn{1}{l}{} & \multicolumn{8}{c||}{\textbf{Text Retrieval}} & \multicolumn{8}{c}{\textbf{Image Retrieval}}

\\ \hline

\multirow{2}{*}{Model} & \multicolumn{2}{c|}{Flickr30K(1K)} & 

\multicolumn{2}{c|}{MS-COCO(1K)} & 

\multicolumn{2}{c|}{\textbf{\centering\begin{tabular}[c]{@{}c@{}}JourneyBench(1K)\\ w/o distractors\end{tabular}}} & 

\multicolumn{2}{c||}{\textbf{\centering\begin{tabular}[c]{@{}c@{}}JourneyBench(1K)\\ w/ distractors\end{tabular}}} & 

\multicolumn{2}{c|}{Flickr30K(1K)} & 

\multicolumn{2}{c|}{MS-COCO(1K)} & 

\multicolumn{2}{c|}{\textbf{\centering\begin{tabular}[c]{@{}c@{}}JourneyBench(1K)\\ w/o distractors\end{tabular}}} &

\multicolumn{2}{c}{\textbf{\centering\begin{tabular}[c]{@{}c@{}}JourneyBench(1K)\\ w/ distractors\end{tabular}}} \\ \cline{2-17} 

 & R@1 & \multicolumn{1}{c|}{R@5} & R@1 & \multicolumn{1}{c|}{R@5} & \hspace{.5em} R@1 & \multicolumn{1}{c|}{ \hspace{.5em} R@5} & \hspace{.5em} R@1 & \multicolumn{1}{c||}{ \hspace{.5em} R@5} & R@1 & \multicolumn{1}{c|}{R@5} & R@1 & \multicolumn{1}{c|}{R@5} & \hspace{.5em} R@1 & \multicolumn{1}{c|}{ \hspace{.5em} R@5} & \hspace{.5em} R@1 & \hspace{.5em} R@5 \\ \hline

ALBEF-210M~\cite{albef}           & 88.50 & \multicolumn{1}{c|}{98.50}                     & 89.10 & \multicolumn{1}{c|}{98.30}                      & \hspace{.5em} 72.30 & \multicolumn{1}{c|}{\hspace{.5em}86.10}            & \hspace{.5em}65.36 & \hspace{.5em}83.75             & 75.90 & \multicolumn{1}{c|}{92.60}                    & 72.28 & \multicolumn{1}{c|}{94.18} & \hspace{.5em}66.12 & \multicolumn{1}{c|}{\hspace{.5em}88.65} & \hspace{.5em}50.02 & \hspace{.5em}75.46 \\

CLIP-430M~\cite{clip}             & 85.30 & \multicolumn{1}{c|}{97.90}                     & 75.60 & \multicolumn{1}{c|}{93.20}                      & \hspace{.5em}70.60 & \multicolumn{1}{c|}{\hspace{.5em}85.70}            & \hspace{.5em}60.80 & \hspace{.5em}83.30             & 64.90 & \multicolumn{1}{c|}{87.20}                    & 54.50 & \multicolumn{1}{c|}{81.80} & \hspace{.5em}66.80 & \multicolumn{1}{c|}{\hspace{.5em}88.80} & \hspace{.5em}51.20 & \hspace{.5em}76.50 \\
 
$\text{X}^2$-VLM-Large-590M~\cite{x2vlm}   & \textbf{98.80} & \multicolumn{1}{c|}{\textbf{100.00}}  & \textbf{93.60} & \multicolumn{1}{c|}{\textbf{99.50}}    & \hspace{.5em}\underline{78.54} & \multicolumn{1}{c|}{\hspace{.5em}92.78}            & \hspace{.5em}64.97 & \hspace{.5em}\textbf{90.47}    & \textbf{91.80} & \multicolumn{1}{c|}{\textbf{98.60}}  & \textbf{83.32} & \multicolumn{1}{c|}{\textbf{96.86}} & \hspace{.5em}75.04 & \multicolumn{1}{c|}{\hspace{.5em}93.16} & \hspace{.5em}61.02 & \hspace{.5em}\textbf{85.00} \\

BEiT3-674M~\cite{beit3}           & 89.50 & \multicolumn{1}{c|}{98.80}                     & 81.10 & \multicolumn{1}{c|}{96.60}                      & \hspace{.5em}74.10 & \multicolumn{1}{c|}{\textbf{\hspace{.5em}87.80}}   & \hspace{.5em}65.90 & \hspace{.5em}86.10             & 75.94 & \multicolumn{1}{c|}{93.34}                    & 66.40 & \multicolumn{1}{c|}{89.50} & \hspace{.5em}68.00 & \multicolumn{1}{c|}{\hspace{.5em}90.30} & \hspace{.5em}56.20 & \hspace{.5em}79.90 \\

BLIP2-12B~\cite{blip2}            & 92.80 & \multicolumn{1}{c|}{\underline{99.90}}                     & \underline{91.30} & \multicolumn{1}{c|}{\underline{99.10}}                      & \hspace{.5em}\textbf{81.29} & \multicolumn{1}{c|}{\hspace{.5em}\underline{95.17}}   & \hspace{.5em}63.78 & \hspace{.5em}\underline{87.76}             & \underline{89.70} & \multicolumn{1}{c|}{\underline{98.10}}                    & \underline{78.78} & \multicolumn{1}{c|}{\underline{94.92}} & \hspace{.5em}75.77 & \multicolumn{1}{c|}{\hspace{.5em}91.66} & \hspace{.5em}59.97 & \hspace{.5em}82.48 \\

OpenCLIP-CoCa-13B~\cite{coca}     & 92.50 & \multicolumn{1}{c|}{99.50}                     & 75.89 & \multicolumn{1}{c|}{93.63}                      & \hspace{.5em}70.43 & \multicolumn{1}{c|}{\hspace{.5em}85.41}            & \hspace{.5em}60.04 & \hspace{.5em}83.32             & 80.40 & \multicolumn{1}{c|}{95.70}                    & 59.30 & \multicolumn{1}{c|}{85.51} & \hspace{.5em}65.83 & \multicolumn{1}{c|}{\hspace{.5em}86.66} & \hspace{.5em}48.70 & \hspace{.5em}72.56 \\

InternVL-C-13B~\cite{Internvl}    & 94.70 & \multicolumn{1}{c|}{99.60}                     & 85.34 & \multicolumn{1}{c|}{96.86}                      & \hspace{.5em}78.22 & \multicolumn{1}{c|}{\hspace{.5em}89.21}            & \hspace{.5em}\textbf{67.73}            & \hspace{.5em}86.41 & 81.70 & \multicolumn{1}{c|}{96.00}            & 71.43 & \multicolumn{1}{c|}{91.50} & \hspace{.5em}\underline{75.84} & \multicolumn{1}{c|}{\hspace{.5em}\underline{93.34}} & \hspace{.5em}\underline{62.29} & \hspace{.5em}83.44 \\

InternVL-G-14B~\cite{Internvl}    & \underline{95.70} & \multicolumn{1}{c|}{99.70}                     & 87.58 & \multicolumn{1}{c|}{97.64}                      & \hspace{.5em}78.52 & \multicolumn{1}{c|}{\hspace{.5em}89.81}            & \hspace{.5em}\underline{67.53} & \hspace{.5em}86.51             & 85.00 & \multicolumn{1}{c|}{97.00}                    & 75.64 & \multicolumn{1}{c|}{93.77} & \hspace{.5em}\textbf{76.80} & \multicolumn{1}{c|}{\hspace{.5em}\textbf{93.80}} & \hspace{.5em}\textbf{63.71} & \hspace{.5em}\underline{84.84}  \\  \hline \hline

\end{tabular}%
}
\caption{\textbf{Zero-shot Evaluation of Cross-modal Retrieval.} The best and second-best results are bolded and underlined. The performance of baseline models on JourneyBench without distractors is comparable to that of existing cross-modal retrieval tasks of similar scale, indicating their generalizability to generated images. However, there is a notable decline in performance when distractors are added, highlighting the critical role of sample-specific distractors in enhancing the challenge of the tasks. Additional results available in appendix.}
\label{tab:crossmodal-retrieval-result}

\end{table*}

\subsection{Quantitative Analysis}

\begin{figure*}[ht!]
    \centering
    \captionsetup{skip=5pt}
    \includegraphics[width=\textwidth,keepaspectratio]{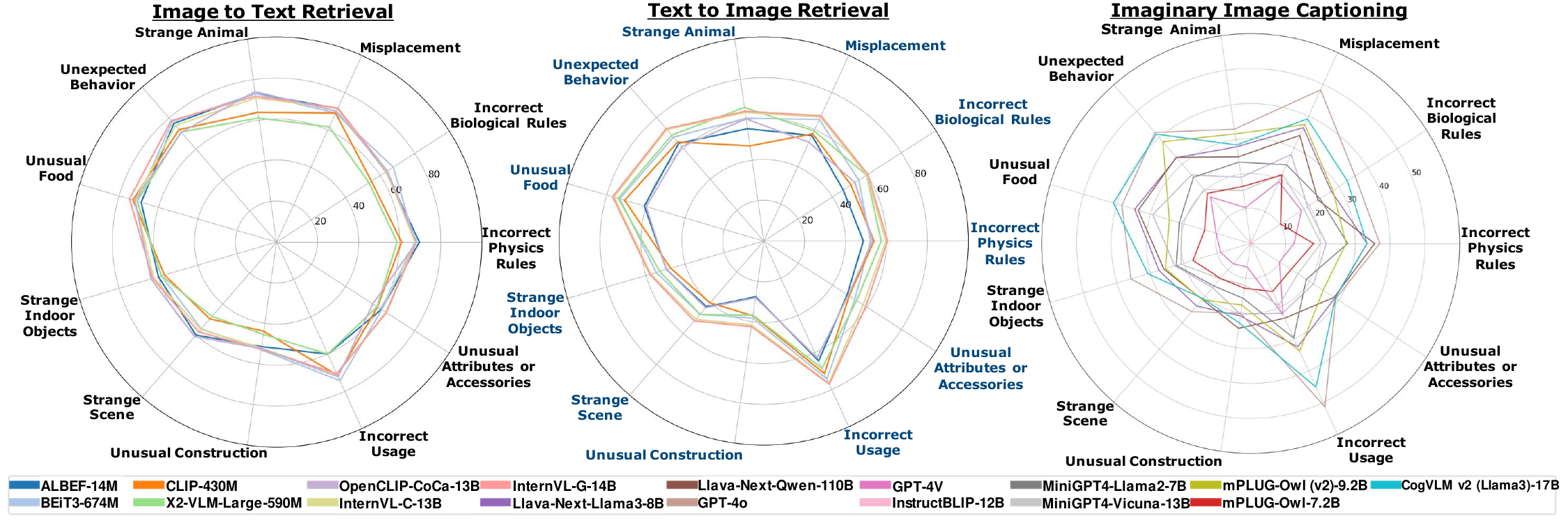}
  \caption{{\textbf{Zero-shot Evaluation on Fine-grained Categories of Retrieval and Captioning.} I2T (left), T2I (center), and Imaginary Image Captioning (right) are measured by Recall@1, Recall@1, and CIDEr respectively. Models particularly struggle with ``Unusual Construction" subcategory.}}
  \label{fig:cr-captioning-across-categories}
  \vspace{-2.5mm}
\end{figure*}

We experimented with various SOTA models on our newly introduced JourneyBench datasets with a range of different experiments, including cross-modal fine-grained retrieval, imaginary image captioning, and multimodal chain-of-thought and multi-image VQA.
 
\textbf{Models struggle with differentiating fine-grained visual details.} We selected a diverse set of models that have previously exhibited strong performance on established cross-modal retrieval datasets \cite{flickr, mscoco, eccv-caption}. Table~\ref{tab:crossmodal-retrieval-result} presents the results of existing SOTA retrieval models on these datasets and our fine-grained cross-modal retrieval dataset.
Among these models, InternVL \cite{Internvl} and BLIP2 \cite{blip2} achieve the highest R@1 score of 67.63\% and 81.29\% for text retrieval with and without distractors, respectively. Regarding image retrieval, with and without distractors, InternVL-G-14B \cite{Internvl} achieved the highest R@1 scores. However, as depicted in Figure~\ref{fig:cr-captioning-across-categories}, the performance of these models on our dataset reveals significant challenges and limitations, with the majority of scores clustered around 60\% and failing to surpass the 80\% mark across all categories. 

The lower recall scores in JourneyBench compared to MS-COCO \cite{mscoco} and Flickr30k \cite{flickr} demonstrate that models encounter greater challenges in retrieving text and images from our dataset. For instance, the highest R@1 performance for text retrieval in MS-COCO-1k is 93.6\%, whereas in JourneyBench with and without distractors, it was only 70.1\% and 81.29\%, respectively. Similarly, for image retrieval, the highest R@1 score on MS-COCO-1k is 83.32\%, which is notably higher than the 76.8\% and 63.71\% scores in our dataset. This disparity highlights the models' struggle in differentiating fine-grained visual and textual details, especially with sample-specific distractors in JourneyBench. 
The varying performance gaps across categories suggest that certain types of image-text relationships are more challenging to capture and align, with categories like "Unusual Construction" and "Strange Scene" 
requiring more sophisticated understanding and reasoning abilities to bridge the semantic gap between the visual and textual modalities.

\begin{wraptable}{l}{0.5\textwidth}

\centering
\resizebox{0.5\textwidth}{!}{%
\begin{tabular}{l|c|c|c|c} \hline \hline
Model                    & BLEU1-4         & CIDEr        & METEOR       & Rouge        \\ \hline
MiniGPT4-Lama2-7B~\cite{minigpt}      & 19.60	& 20.91	&  18.07	&  28.76 \\
mPLUG-Owl-7.2B~\cite{mplug}       & 19.53	 & 14.68	& 19.32	& 27.66 \\
LLaVA-Next-Llama3-8B~\cite{llavanext}     & 20.01	 & 28.69	& 15.01	& 26.38 \\
mPLUG-Owl (v2)-9.2B~\cite{mplug}    & \textbf{24.31}	& 26.74	&\underline{20.51}	&\underline{30.97}\\
Blip-2-12B~\cite{blip2} &  17.75	&26.00	&\textbf{22.00}	&\textbf{37.00}  \\
InstructBLIP-12B~\cite{instructblip} & 10.23	 &00.46	&17.19	&19.51 \\
OpenCLIP-CoCa-13B ~\cite{coca}   &18.79	&21.59	&12.02	&24.40\\
MiniGPT4-Vicuna-13B~\cite{minigpt}    & 12.79	&16.21	&17.10	&24.51 \\
CogVLM v2 (lama3)-17B~\cite{cogvlm}    &\underline{21.86}	&\underline{30.31}	&18.63	&28.67 \\ 
LLaVA-Next-Qwen110B~\cite{llavanext}       & 19.73	& 27.18	& 14.96	& 26.61 \\
GPT-4o                   &\underline{21.86}	&\textbf{32.56}	&18.56	&28.37 \\
GPT-4V                   & 17.36	 &11.24	&19.47	&26.75 \\ \hline \hline
\end{tabular}
}
\caption{{Zero-shot Evaluation on Imaginary Image Captioning.} The best and second-best results are bolded and underlined. The low scores on the metrics indicate the baselines struggle to describe imaginary images.}
\label{tab:ugc_result}

\vspace{-2em}

\end{wraptable}\textbf{Models are not used to imaginary visual scenarios. }
We conducted experiments that included various SOTA models for visual understanding, such as LLaVA-NeXT \cite{llavanext}, MiniGPT-4 \cite{minigpt}, mPlug-Owl \cite{mplug, mplugowl2},  GPT-4o, etc. for the captioning task. In Table \ref{tab:ugc_result} and Figure \ref{fig:cr-captioning-across-categories}, most of the models performed poorly on JourneyBench compared to their performance on other captioning datasets \cite{flickr, mscoco, eccv-caption}, with the majority of the models achieving CIDEr scores less than 30.

\begin{wrapfigure}{r}{0.6\textwidth}
  \begin{center}
  \vspace{-1.5em}
    \centering
    \captionsetup{skip=5pt}
\includegraphics[width=0.6\textwidth, keepaspectratio]{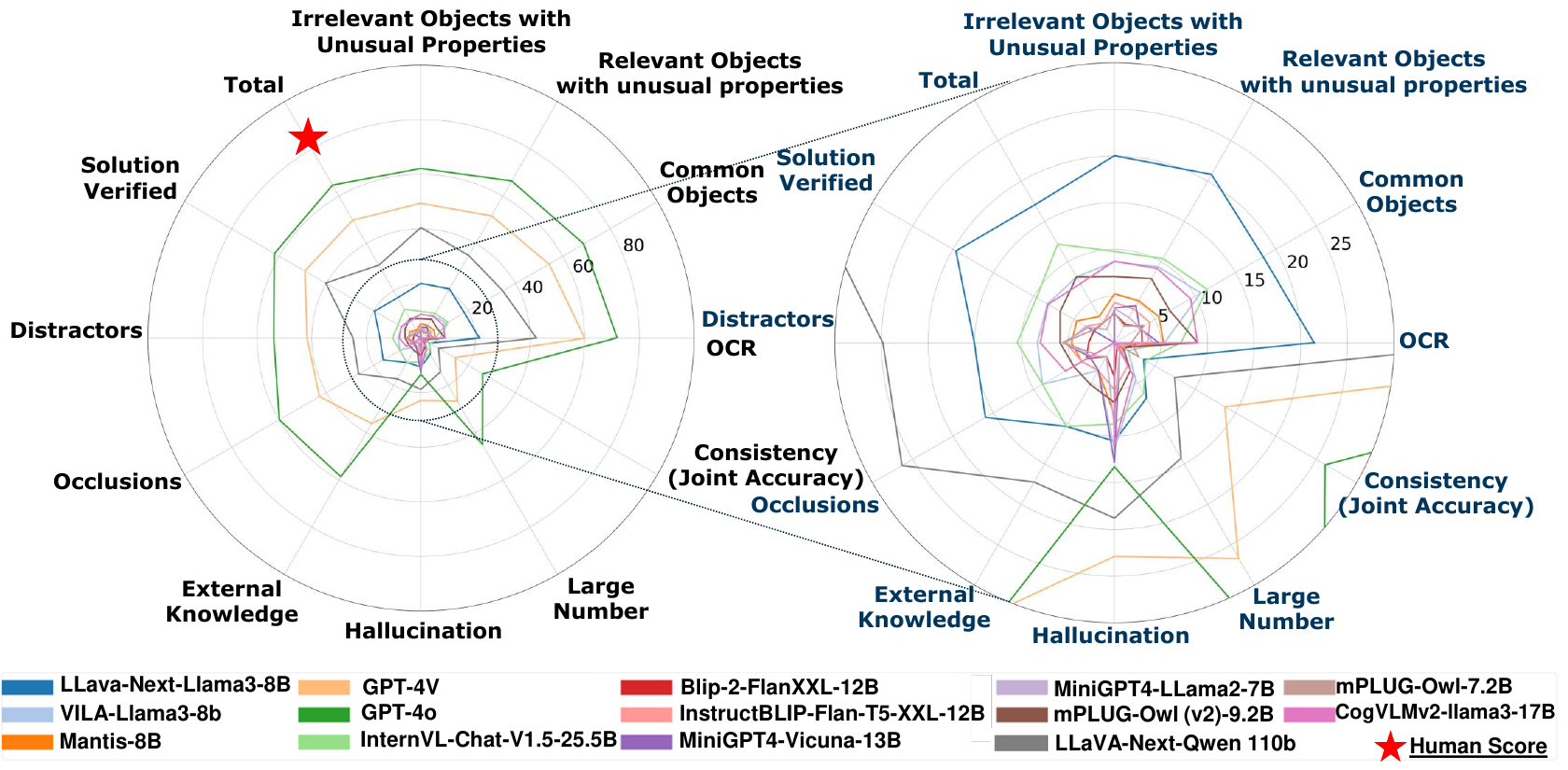}
  \end{center}
  \vspace{-1em}
  \caption{{\textbf{Zero-shot Evaluation on Fine-grained Categories of MCOT.} Models struggle to get high accuracy in all categories, especially for image-question pairs with hallucinations or with large numbers of objects.}}
    \label{fig:mcot_across_categories}
  \vspace{-2em}
\end{wrapfigure}

\textbf{Co-referencing across modalities is challenging in arithmetic reasoning.}
Figure~\ref{fig:mcot_across_categories} illustrates the performance of SOTA methods across fine-grained categories of the JourneyBench MCOT dataset. 
Our complementary MCOT task proves to be highly challenging, with GPT-4o achieving only 62.18$\%$ accuracy. Most other models, except GPTs and LLaVAs, score below 10$\%$. Notably, GPT-4V and GPT-4o struggle with consistency, hallucination, and co-referencing in visual contexts with numerous objects. Additionally, smaller VLMs also find it difficult to utilize external knowledge when solving MCOT questions. 

 To demonstrate the complementary nature of our image-question pairs in MCOT, we tested a language-only GPT-4o model on our dataset, which resulted in just 16.64$\%$ accuracy. In contrast, 
language-only GPT-4o achieved 83.9$\%$ on ScienceQA \cite{scienceqa}. This significant difference underscores the importance of complementary visual and textual information in multimodal reasoning tasks. The red star in Figure \ref{fig:mcot_across_categories} indicates human performance at 84\%, suggesting that there is still significant room for improvement even for the SOTA LLMs.

\begin{wraptable}{r}{0.6\textwidth}

\vspace{-1em}

\centering

\resizebox{0.6\textwidth}{!}{
     \begin{tabular}{l|l|l|m{1 cm}|m{1 cm}|m{1 cm}|c|l}  \hline \hline
\multirow{3}{*}{Model} & \multicolumn{6}{c|}{Multi-Image VQA} & \multirow{3}{*}{\makecell{Mantis\\Eval}} \\ \cline{2-7}
   &   \multirow{2}{*}{All} &  \multicolumn{4}{c|}{MMCOT} & \multirow{2}{*}{\makecell{Cause \\and \\ Effect}} &  \\ \cline{3-6}
   &   & \multicolumn{1}{l|}{All} &\multicolumn{1}{l|}{\makecell{Arithmetic \\ Reasoning}} & \multicolumn{1}{l|}{\makecell{External\\Knowledge}} & \multicolumn{1}{l|}{\makecell{Solution\\Verification}} &   & \\ \hline
VILA-8B~\cite{vila}            &    24.20   & 6.14  & 3.73  & 8.65  & 3.77 & 53.92 & 51.15 \\
Idefics2-8B~\cite{idefics2}        & 27.82 & 6.61  & 2.81  & 10.57 & 4.95 & 65.03 & 48.85 \\
Mantis-Idefics2-8B~\cite{mantis} & 19.90 & 3.30  & 3.71  & 2.88  & 7.26 & 49.02 & 57.14 \\
Mantis-SigLIP-8B~\cite{mantis}   & 23.29 & 4.72  & 5.98  & 3.41  & 7.82 & 55.88 & 59.45 \\
GPT-4V~              & \underline{48.70} & \underline{32.54} & \underline{32.88} & \textbf{32.2}  & \underline{36.31} & \underline{77.06} & \underline{62.67} \\
GPT-4o~              & \textbf{56.39} & \textbf{41.03} & \textbf{52.04} & \underline{29.61} & \textbf{43.39} & \textbf{83.33} &  \textbf{73.42}   \\ 
\hline
Human&    78.90 &    71.40         &  86.00     &     55.80       &    -  &   92.00    &  - \\
\hline \hline
\end{tabular}
}
\caption{\textbf{Zero-shot Evaluation on Multi-Image Visual Reasoning.} The best and second-best results are bolded and underlined. Models like GPT-4o perform worse on our Multi-image VQA or MMCOT than on Mantis-Eval. Note that most models on Cause and Effect - being a binary-choice question - have an accuracy of nearly random guessing. }
\label{tab:multi-image mcot}

\vspace{-1em}

\end{wraptable}

\textbf{Co-referencing across multiple images is extremely challenging.}
Table~\ref{tab:multi-image mcot} presents the performance of different SOTA VLMs on our proposed multi-image VQA dataset across various categories, as well as on the Mantis-Eval dataset. 
Overall, models encountered greater challenges in co-referencing across multiple images in JourneyBench, with low scores in the range of 39.04\% $\pm$ 18.85\%. Especially concerning MMCOT VQA, performance is even lower in the range of 23.58\% $\pm$ 19.81\% across different SOTA VLMs. 
Meanwhile, all the models achieved much higher accuracy scores in the range of 61.13\% $\pm$ 12.29\% on the Mantis-Eval dataset. For instance, GPT-4o achieved an accuracy of 73.42\% on the Mantis-Eval dataset, which is approximately 32 $\%$ and 17\% higher than its performance, 41.03$\%$ on our MMCOT and 56.39$\%$ on our multi-image VQA. Similar to our MCOT task, we also conduct a human evaluation to obtain an estimation of the expected maximum performance. As shown in the figure, the arithmetic reason is similar to MCOT, suggesting humans are indifferent to multiple images. However, since we restrict access to the internet during the human test, the low external knowledge result causes a significant drawback to the overall score.

\begin{wrapfigure}{l}{0.60\textwidth}
    \vspace{-1.3em}

  \centering
\includegraphics[width=0.60\textwidth, keepaspectratio]{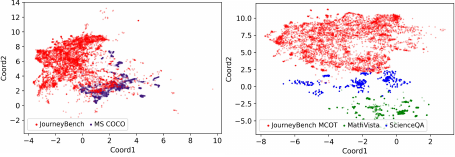}
  \caption{\textbf{Low-dimensional Representation of JourneyBench, MS COCO, MathVista, and ScienceQA Images.} JourneyBench shows a more diverse distribution.}
  \label{fig:visual_data_distribution_1}
  \vspace{-1.1em}
\end{wrapfigure}

\subsection{Qualitative Analysis}
\label{sec:qualitative_qnalysis}
\textbf{Image Diversity Visualization.} 
Figure \ref{fig:visual_data_distribution_1} shows the result of dimension reduction using UMAP \cite{umap_2018} on CLIP's embedding space, sampling an equal number of images from each dataset. 
In the top figure, JourneyBench's distribution is not only more expansive but also encompasses the majority of COCO's data distribution, suggesting a richer semantic diversity. 
The bottom figure shows JourneyBench's MCOT images have a similarly diverse distribution. Compared to existing MCOT benchmarks like MathVista \cite{mathvista}, and ScienceQA \cite{scienceqa}, JourneyBench MCOT displays significantly greater diversity. Despite sampling an equal number of images from each dataset, JourneyBench appears more populated in the graph. This is because images in MathVista and ScienceQA are often very similar, such as maps, tables, and illustrations that change only slightly, resulting in densely overlapping data points in the UMAP visualization.

\vspace{-1em}

\section{Conclusion} 
We introduce JourneyBench, a new benchmark that tests models' understanding of unusual or fictional images across various tasks, including multimodal chain-of-thought, multi-image VQA, image captioning, visual question answering, and cross-modal retrieval. JourneyBench's tasks consistently yield lower evaluation scores from all tested baseline models, underscoring the challenges posed by its unusual or fictional image subjects, strategically designed distractors, hallucination-inducing questions, and questions that require cross-modal referencing. This makes JourneyBench an ideal tool for assessing the capabilities of advanced MM-LLMs, pushing the boundaries of what these models can understand and interpret.

\bibliographystyle{splncs04}
\bibliography{neurips2024}

\section*{Checklist}


\begin{enumerate}

\item For all authors...
\begin{enumerate}
  \item Do the main claims made in the abstract and introduction accurately reflect the paper's contributions and scope?
    \answerYes{}
  \item Did you describe the limitations of your work?
    \answerYes{See supplemental materials}
  \item Did you discuss any potential negative societal impacts of your work?
    \answerYes{See supplementary materials}
  \item Have you read the ethics review guidelines and ensured that your paper conforms to them?
    \answerYes{}
\end{enumerate}

\item If you are including theoretical results...
\begin{enumerate}
  \item Did you state the full set of assumptions of all theoretical results?
    \answerNA{}
	\item Did you include complete proofs of all theoretical results?
    \answerNA{}
\end{enumerate}

\item If you ran experiments (e.g. for benchmarks)...
\begin{enumerate}
  \item Did you include the code, data, and instructions needed to reproduce the main experimental results (either in the supplemental material or as a URL)?
    \answerYes{See supplemental materials}
  \item Did you specify all the training details (e.g., data splits, hyperparameters, how they were chosen)?
    \answerYes{See supplemental materials}
	\item Did you report error bars (e.g., with respect to the random seed after running experiments multiple times)?
    \answerNo{For the experiment types that we run, it is not customary to report error bars. Further, there is minimum randomness, as we use off-the-shelf pretrained models with fixed checkpoints.}
	\item Did you include the total amount of compute and the type of resources used (e.g., type of GPUs, internal cluster, or cloud provider)?
    \answerYes{See supplemental materials}
\end{enumerate}

\item If you are using existing assets (e.g., code, data, models) or curating/releasing new assets...
\begin{enumerate}
  \item If your work uses existing assets, did you cite the creators?
    \answerYes{}
  \item Did you mention the license of the assets?
    \answerYes{}
  \item Did you include any new assets either in the supplemental material or as a URL?
    \answerYes{See supplemental materials}
  \item Did you discuss whether and how consent was obtained from people whose data you're using/curating?
    \answerYes{See supplemental materials}
  \item Did you discuss whether the data you are using/curating contains personally identifiable information or offensive content?
    \answerYes{See supplemental materials}
\end{enumerate}

\item If you used crowdsourcing or conducted research with human subjects...
\begin{enumerate}
  \item Did you include the full text of instructions given to participants and screenshots, if applicable?
    \answerYes{See supplemental materials}
  \item Did you describe any potential participant risks, with links to Institutional Review Board (IRB) approvals, if applicable?
    \answerNo{Our dataset consists of publicly available data. No interaction with human subjects or personally identifiable information  with human subjects is collected, so we do not need IRB approval.}
  \item Did you include the estimated hourly wage paid to participants and the total amount spent on participant compensation?
    \answerYes{See supplemental materials}
\end{enumerate}

\end{enumerate}

\section*{Appendix}
\appendix

\tableofcontents

\clearpage 

\section{Project Page and Dataset Access}
You can directly access the data via \url{https://journeybench.github.io/}

\section{Code Access}
You can directly access the code via \url{https://github.com/JourneyBench/JourneyBench}

\section{Evaluation Procedure}
\subsection{Inference Prompts}
In this section, we list the inference prompts for models to generate responses across JourneyBench tasks, including MCOT, Multi-image MCOT (MMCOT), Multi-image Case and Effect, Imaginary Image Captioning and HaloQuest (VQA with hallucination triggers).

\textbf{MCOT}
\begin{verbatim}
"""
You will be provided with an image and a mathematical question. 
You need to solve the question with the information from the image.

Question: {$question}
"""
\end{verbatim}

\textbf{Multi-image MCOT}
\begin{verbatim}
"""
You will be provided with two images and a mathematical question. 
You need to solve the question with the information from the images.

Question: {$question}
"""
\end{verbatim}
\textbf{Multi-image Cause and Effect}
\begin{verbatim}
"""
You will be provided with two images <image1> and <image2> and a question 
querying the causal relationship between the concepts described in the 
images or text. 
Your final answer must be one of <image1> or <image2>.

Question: {$question}
"""
\end{verbatim}

\textbf{Imaginary Image Captioning}
\begin{verbatim}
"""
Describe the unusual feature of the image in a single sentence.
"""
\end{verbatim}

\textbf{HaloQuest (VQA with Hallucination Triggers)}

We use the default VQA prompt of each model. If no default VQA prompt is provided, we use the following prompt:
\begin{verbatim}
"""
Question: {$question} Answer: 
"""
\end{verbatim}

\subsection{MCOT/MMCOT Answer Extraction $\&$ Verification}
VLMs can produce not only the numerical answer but also the mathematical reasoning steps taken to arrive at the answer. 
Because the answer format can vary (e.g.~1/2=0.5=4/8), verifying the answer accuracy requires extra steps.
Other works, for example, ScienceQA \cite{scienceqa} use a regular expression to extract the produced answer from ChatGPT, since it consists of only multiple-choice questions. 
However, due to the nature of MCOT and MMCOT, distinguishing the final numerical answer from other numbers in the calculation steps can be challenging. Further, even if one prompts the VLM to produce the answer in the correct format (e.g.~always express the answer in decimal on the last line), models may sometimes fail to follow the instruction or may contain a variable number of decimal points. 
Thus, we use Meta-Llama3-8B-Instruct \cite{llama-3} to first extract the answer using the prompt: 
\begin{verbatim}
"""
Question: {$question} 

Solution:{$reasoning_steps}. 

The solution is generated by an AI model. 
Identify and extract the final numeric answer from the solution. 
If the answer is not explicitly stated as a number, infer it if possible. 
If no numeric answer can be determined, respond with ‘unknown’. 
Output only the numeric answer or ‘unknown’.
"""
\end{verbatim}
Once the final numerical solution has been extracted, we then use the same model to verify the answer using the prompt: 
\begin{verbatim}
"""
Question: {$question} 

Predicted Answer: {$predicted_answer} 

Ground Truth Answer: {$ground_truth_answer} 

Does the predicted answer match the ground truth answer and directly address 
the question? 
If the absolute difference between their values is within 0.1, answer ‘yes’; 
otherwise, answer ‘no’. Respond only with ‘yes’ or ‘no’.
"""
\end{verbatim}

The verification results are reported in the form of "yes" and "no". From this, we can calculate the accuracy of the model's answers (we call this step ``Answer Verification'').

\subsection{MCOT/MMCOT Solution Verification}
We next seek to determine whether the model follows the correct logic and steps when solving the problem. We also provide step-by-step solutions in our MCOT and MMCOT annotation. 
To perform solution verification, we employ a Meta-Llama3-70B-Instruct \cite{llama-3}. Essentially, we ask the language model to compare the generated solution with the ground truth provided solution and to determine whether the predicted reasoning steps follow the same approach and lead to the correct solution. 
We prompt Llama3-70B-Instruct using the prompt: 

\begin{verbatim}
"""
Question: {$question}

Predicted Reasoning Steps: {$predicted_reasoning_steps} 

Ground Truth Reasoning Steps: {$ground_truth_reasoning_steps} 

Do the predicted reasoning steps follow the same approach as the ground
truth reasoning steps and lead to the correct solution? 

Respond with 'yes' if they match, or 'no' if they differ significantly or
lead to an incorrect solution.
"""
\end{verbatim}

\subsection{HaloQuest Answer Evaluation}
HaloQuest is an open-ended visual question answering dataset focusing on testing VLMs with hallucination triggers. Unlike our MCOT task, HaloQuest does not ask questions requiring mathematical problem solving skills, but instead asks general questions about the image. HaloQuest features questions designed to trigger models to provide a hallucination response via false premise questions (question assumes something not true in the image), visually challenging questions (answering the question requires visual aspects of the image that are hard to see), and questions with insufficient context to answer (asking about something not visible in the image). 
HaloQuest is a generalizable dataset for future VLMs as it allows free-form answer verification, rather than requiring models to answer multiple choice questions. 
We follow \cite{haloquest} to conduct the answer extraction and verification process. 
To make the evaluation process more consistent across the five tasks in JourneyBench, we also adopt Llama3-8B-Instruct to first extract and then verify the answer based on the raw responses, ground-truth answers, and questions. Specifically, we used the prompt to conduct answer extraction.

\begin{verbatim}
"""
Answer extractor. 

Here is my question: {$question} 

Here is the response: {$response} 

Can you help me extract the answer from the response to my question? Your 
extracted answer should be short in one sentence. 
"""
\end{verbatim}

In addition, we used the prompt below to conduct answer verification using the LLM. That is, we had Llama3-8B-Instruct serve as a judge by giving it the ground truth answer and the predicted answer and asking it to determine if the predicted answer is correct given the ground truth.

\begin{verbatim}
"""
Answer verifier.

Your task is to determine if the model response is correct given the question
and ground truth response. Ensure that the model response is by the question. 

If the question asks about a detail of an element that is not present in the
image, A prediction of "yes", "no" or "nothing" should be considered incorrect
because it inaccurately suggests that the element is presented in the image. The 
correct prediction in such cases should acknowledge the absence of the element 
in question by stating the element is not present.

If the question is about counting, then the prediction is correct only if it 
matches the ground truth counts exactly.

question = {$question},
model_response = {$model_response}
groundtruth_response = {$groundtruth_response}

Please only output 'Yes' or 'No.'
"""
\end{verbatim}
\section{Detailed Experiment Results}
\begin{table}[!t]
\captionsetup{skip=5pt}
\centering
\fontsize{8}{9}
\selectfont
\setlength{\tabcolsep}{3pt}
\renewcommand{\arraystretch}{1.4}
\resizebox{\textwidth}{!}{%
\begin{tabular}{l|c|c|c|c|c|c|c} \hline \hline
Model&Mean Rank&MCOT&Multi-image VQA&Captioning(C)&Text R@1&Image R@1&VQA + Hall. Trig.\\\hline
GPT-4o                      &\textbf{2.83}&\textbf{62.18}&\textbf{56.39}&32.56&&&47.86\\
LLaVA-Next-Qwen 110B        &3.16&40.43&    &27.18&&&\textbf{60.03}\\
LLaVA-Next-Llama3-8B        &3.16&20.03&    &28.69&&&39.63\\
VILA-Llama3-8B              &4.0&8.66 &24.19&\textbf{33.79}&&&21.38\\
Mantis-Idefics2-8B          &4.5&5.25 &19.90&33.34&&&24.87\\
GPT-4V                      &4.67&49.34&48.7 &11.24&&&44.73\\
BEiT3 Large-0.7B            &5.16&4.10 &    &30.90&\textbf{65.90}&56.20&34.21\\
Blip-2-FlanXXL-12B          &6.33&3.13 &    &26.00&63.78&\textbf{59.97}&20.56\\
CogVLM v2 (Llama3)-19B      &8.33&8.73 &    &30.31&     &              &38.48\\
InstructBLIP-Flan-T5-XXL-12B&8.5&4.31 &    &0.46 &&&24.83\\
MiniGPT4-Vicuna13B          &9.0&3.73 &    &16.21&&&23.39\\
mPLUG-Owl v2-9.2B           &9.5&7.07 &    &26.74&&&9.53\\
MiniGPT4-Llama2-7B          &10.0&3.69 &    &20.91&&&15.27\\
mPLUG-Owl-7.2B              &11.83&3.19 &    &14.68&&&8.05\\
\hline 
Human                       &&84.0 &78.9 &85.71&&&84.61\\
Random Basline              &&0    &16.56&    &0.01&0.02&0.83\\\hline
\hline
 
\end{tabular}%
}
\caption{Overview of model performance on all datasets. Captioning scores measured in CIDEr. VQA + Hall. Trig. stands for VQA + Hallucination Triggers (HaloQuest). We calculate mean rank by first ranking the model's performance on each task and taking the mean, blank cells are treated as a score of zero during ranking.}
\label{tab:main_table}
\end{table}
\subsection{Experiment Results Across Five Tasks}
In Table \ref{tab:main_table} we show a comprehensive overview of model performances on all our datasets. Note that in Table \ref{tab:main_table} we only show models that are capable of running on three or more tasks (i.e.~some cross-modal retrieval models can't perform other types of tasks). We observe several surprising findings in JourneyBench. Perhaps one of the most surprising findings is that the LLaVA-Next-Qwen-110b model outperforms GPT-4o and GPT-4V significantly on the HaloQuest benchmark. This shows that GPT is significantly more prone to hallucinations than this open source model. This has implications for downstream applications where hallucination-inducing questions are likely. Users using GPT in such applications should be aware that its performance exhibits significant drops in the presence of such questions.

Note that the random baseline is higher for multi-image VQA due to the inclusion of binary cause $\&$ effect questions as part of this task. For multi-image mathematical reasoning questions, random performance is the same as MCOT.

\begin{table}[!t]
\captionsetup{skip=5pt}
\centering
\selectfont
\setlength{\tabcolsep}{3pt}
\renewcommand{\arraystretch}{1.4}
\resizebox{\textwidth}{!}{%
\begin{tabular}{l|ccccccccccccccc}
\hline 
\hline
\multicolumn{1}{l|}{} & \multicolumn{15}{c}{\textbf{Text Retrieval}} \\ \hline

\multirow{2}{*}{Model} & \multicolumn{3}{c|}{Flicker30K-Full} & \multicolumn{3}{c|}{\begin{tabular}[c]{@{}c@{}}MS-COCO-Full\end{tabular}} & \multicolumn{3}{c|}{\begin{tabular}[c]{@{}c@{}}MS-COCO-1K\end{tabular}} & \multicolumn{3}{c|}{\textbf{\begin{tabular}[c]{@{}c@{}}JourneyBench-1K\\ w/o distractors\end{tabular}}} & \multicolumn{3}{c}{\textbf{\begin{tabular}[c]{@{}c@{}}JourneyBench-1K\\ w/ distractors\end{tabular}}} \\ \cline{2-16}

 & R@1 & R@5 & \multicolumn{1}{c|}{R@10} & R@1 & R@5 & \multicolumn{1}{c|}{R@10} & R@1 & R@5 & \multicolumn{1}{c|}{R@10} & R@1 & R@5 & \multicolumn{1}{c|}{R@10} & R@1 & R@5 & R@10 \\ \hline

ALBEF-210M~\cite{albef} & 88.5 & 98.5 & \multicolumn{1}{c|}{99.2} & 73.96 & 91.8 & \multicolumn{1}{c|}{96.0} & 89.1 & 98.3 & \multicolumn{1}{c|}{99.6} & 72.3 & 86.1 & \multicolumn{1}{c|}{91.78} & 65.36 & 83.75 & 89.13 \\

BEiT3-674M~\cite{beit3} & 89.5 & 98.8 & \multicolumn{1}{c|}{99.4} & 64 & 86.6 & \multicolumn{1}{c|}{92.2} & 81.1 & 96.6 & \multicolumn{1}{c|}{98.8} & 74.1 &87.80 & \multicolumn{1}{c|}{92.70} & 65.9 & 86.1 & 90.9 \\

BLIP2-12B~\cite{blip2} & 92.8 & 99.9 & \multicolumn{1}{c|}{99.9} & 80.1 & 94.8 & \multicolumn{1}{c|}{97.9} & 91.3 & 99.1 & \multicolumn{1}{c|}{99.6} & \textbf{81.29} & \textbf{95.17} & \multicolumn{1}{c|}{\textbf{97.28}} & 63.78 & 87.76 & 92.46 \\

CLIP-430M~\cite{clip} & 85.3 & 97.9 & \multicolumn{1}{c|}{99.1} & 58.4 & 81.5 & \multicolumn{1}{c|}{88.1} & 75.6 & 93.2 & \multicolumn{1}{c|}{97.5} & 70.6 & 85.7 & \multicolumn{1}{c|}{91} & 60.8 & 83.3 & 88.5 \\

X$^2$\-VLM-Large-590M~\cite{x2vlm} & \textbf{98.8} & \textbf{100} & \multicolumn{1}{c|}{\textbf{100}} & \textbf{84.4} & \textbf{96.5} & \multicolumn{1}{c|}{\textbf{98.5}} & \textbf{93.6} & \textbf{99.5} & \multicolumn{1}{c|}{\textbf{99.9}} & 78.54 & 92.78 & \multicolumn{1}{c|}{96.15} & 64.97 & \textbf{90.47} & \textbf{94.8} \\

InternVL-C-13B~\cite{Internvl} & 94.7 & 99.6 & \multicolumn{1}{c|}{99.9} & 74.9 & 91.3 & \multicolumn{1}{c|}{95.2} & 85.34 & 96.86 & \multicolumn{1}{c|}{98.84} & 78.22 & 89.21 & \multicolumn{1}{c|}{93.61} & \textbf{67.73} & 86.41 & 91.91 \\

InternVL-G-14B~\cite{Internvl} & 95.7 & 99.7 & \multicolumn{1}{c|}{99.9} & 74.9 & 91.3 & \multicolumn{1}{c|}{95.2} & 87.58 & 97.64 & \multicolumn{1}{c|}{99.28} & 78.52 & 89.81 & \multicolumn{1}{c|}{94.21} & 67.53 & 86.51 & 92.61 \\

OpenCLIP-CoCa-13B~\cite{coca} & 92.5 & 99.5 & \multicolumn{1}{c|}{99.9} & 66.3 & 86.2 & \multicolumn{1}{c|}{91.8} &75.89	&93.63 & \multicolumn{1}{c|}{97.15 } & 70.43 & 85.41 & \multicolumn{1}{c|}{89.61} & 60.04 & 83.32 & 87.91 \\ \hline 

\multicolumn{1}{l|}{} & \multicolumn{15}{c}{\textbf{Image Retrieval}} \\ \hline

ALBEF-210M~\cite{albef} & 75.9 & 92.6 & \multicolumn{1}{c|}{96} & 54 & 78.99 & \multicolumn{1}{c|}{87.18} & 72.28 & 94.18 & \multicolumn{1}{c|}{97.54} & 66.12 & 88.65 & \multicolumn{1}{c|}{92.15} & 50.02 & 75.46 & 82.56 \\

BEiT3-674M~\cite{beit3} & 75.94 & 93.34 & \multicolumn{1}{c|}{96.66} & 48.9 & 73.2 & \multicolumn{1}{c|}{81.8} & 66.4 & 89.5 & \multicolumn{1}{c|}{95.2} & 68 & 90.3 & \multicolumn{1}{c|}{94.1} & 56.2 & 79.9 & 85.7 \\

BLIP2-12B~\cite{blip2} & 89.7 & 98.1 & \multicolumn{1}{c|}{98.9} & 63 & 84.2 & \multicolumn{1}{c|}{90.2} & 78.78 & 94.92 & \multicolumn{1}{c|}{97.74} & 75.77 & 91.66 & \multicolumn{1}{c|}{94.12} & 59.97 & 82.48 & 87.17 \\

CLIP-430M~\cite{clip} & 64.9 & 87.2 & \multicolumn{1}{c|}{92} & 37.8 & 62.4 & \multicolumn{1}{c|}{72.2} & 54.5 & 81.8 & \multicolumn{1}{c|}{91} & 66.8 & 88.8 & \multicolumn{1}{c|}{92.5} & 51.2 & 76.5 & 83.5 \\

X$^2$VLM-Large-590M~\cite{x2vlm} & \textbf{91.8} & \textbf{98.6} & \multicolumn{1}{c|}{\textbf{99.5}} & \textbf{67.7} & \textbf{87.5} & \multicolumn{1}{c|}{\textbf{92.5}} & \textbf{83.32} & \textbf{96.86} & \multicolumn{1}{c|}{\textbf{98.6}} & 75.04 & 93.16 & \multicolumn{1}{c|}{95.9} & 61.02 & 85 & 89.69 \\

InternVL-C-13B~\cite{Internvl} & 81.7 & 96 & \multicolumn{1}{c|}{98.2} & 54.1 & 77.3 & \multicolumn{1}{c|}{84.6} & 71.43 & 91.5 & \multicolumn{1}{c|}{96.28} & 75.84 & 93.34 & \multicolumn{1}{c|}{96.31} & 62.29 & 83.44 & 89.33 \\

InternVL-G-14B~\cite{Internvl} & 85 & 97 & \multicolumn{1}{c|}{98.6} & 58.6 & 81.3 & \multicolumn{1}{c|}{88.0} & 75.64 & 93.77 & \multicolumn{1}{c|}{97.48} & \textbf{76.8} & \textbf{93.8} & \multicolumn{1}{c|}{\textbf{96.4}} & \textbf{63.71} & \textbf{84.84} & \textbf{90.28} \\

OpenCLIP-CoCa-13B~\cite{coca} & 80.4 & 95.7 & \multicolumn{1}{c|}{97.7} & 51.2 & 74.2 & \multicolumn{1}{c|}{82.0} &59.30	&85.51 & \multicolumn{1}{c|}{92.78} & 65.83 & 86.66 & \multicolumn{1}{c|}{91.41} &48.70	&72.56	&80.53  \\ \hline \hline
\end{tabular}%
 }
\caption{Zero-shot evaluation of retrieval tasks on different datasets along with our proposed JourneyBench fine-grained cross-modal retrieval datasets. The best results are highlighted in bold.}
\label{tab:crossmodal-retrieval-result}
\end{table}

\subsection{Detailed Retrieval Results}
In table \ref{tab:crossmodal-retrieval-result} we include detailed cross-modal retrieval results beyond those found in our main text, including R@10 for each dataset. \footnote{The ALBEF and CogVLM v2 parameter sizes in the main paper figure were labeled incorrectly and will be fixed later.}
We observe that of all models, X$^2$VLM-Large-590M performs quite strongly across multiple benchmarks for its size. For example, on FlickR30k, it achieves $98.8$ R@1 for text retrieval and 91.8 for image retrieval, despite being more than ten times smaller than several other worse performing models (e.g.~OpenCLIP-CoCa-13B, InternVL-G-14B). We observe that it also performs extremely competitively across MS-COCO and JourneyBench without distractors. However, in the presence of sample-specific distractors, it performs worse. We observe that all models are relatively close with distractors, e.g.~50s to low 60s for R@1 for image retrieval, and 60s for text retrieval. One observation is that model size may be more significant in more complex retrieval scenarios, with larger models either catching up and outperforming it on JourneyBench with distractors. This might indicate that larger models are better able to distinguish fine-grained details than X$^2$-VLM-Large-590M, which excels at course grained retrieval tasks.

\begin{table}[t]
\captionsetup{skip=5pt}
\centering
\fontsize{8}{9}
\selectfont
\setlength{\tabcolsep}{3pt}
\renewcommand{\arraystretch}{1.4}
\resizebox{0.65\textwidth}{!}{%
     \begin{tabular}{l|c|c|m{1 cm}|m{1 cm}|m{1 cm}|c|l}  \hline \hline
\multirow{3}{*}{Model} & \multicolumn{6}{c|}{Multi-Image VQA} & \multirow{3}{*}{\makecell{Mantis\\Eval}} \\ \cline{2-7}
   &   \multirow{2}{*}{All} &  \multicolumn{4}{c|}{MMCOT} & \multirow{2}{*}{\makecell{Cause \\and \\ Effect}} &  \\ \cline{3-6}
   &   & \multicolumn{1}{l|}{All} &\multicolumn{1}{l|}{\makecell{Arithmetic \\ Reasoning}} & \multicolumn{1}{l|}{\makecell{External\\Knowledge}} & \multicolumn{1}{l|}{\makecell{Solution\\Verification}} &   & \\ \hline
VILA-8B~\cite{vila}            &    24.20   & 6.14  & 3.73  & 8.65  & 3.77 & 53.92 & 51.15 \\
Idefics2-8B~\cite{idefics2}        & 27.82 & 6.61  & 2.81  & 10.57 & 4.95 & 65.03 & 48.85 \\
Mantis-Idefics2-8B~\cite{mantis} & 19.90 & 3.30  & 3.71  & 2.88  & 7.26 & 49.02 & 57.14 \\
Mantis-SigLIP-8B~\cite{mantis}   & 23.29 & 4.72  & 5.98  & 3.41  & 7.82 & 55.88 & 59.45 \\
GPT-4V~              & 48.70 & 32.54 & 32.88 & \textbf{32.2}  & 36.31 & 77.06 & 62.67 \\
GPT-4o~              & \textbf{56.39} & \textbf{41.03} & \textbf{52.04} & 29.61 & \textbf{43.39} & \textbf{83.33} &  \textbf{73.42}   \\ 
\hline
Human&    78.90 &    71.40         &  86.00     &     55.80       &    -  &   92.00    &  - \\
Human+Internet&    86.39 &    83.2         &  86.00     &     78.9       &    -  &   92.00    &  - \\
\hline \hline
\end{tabular}
}
\caption{Zero-shot Evaluation on Multi-Image Visual Reasoning. }
\label{tab:multi-image mcot_appendix}
\end{table}
\subsection{Additional Multi-image VQA Results}
In Table \ref{tab:multi-image mcot_appendix} we provide additional analysis of human performance on multi-image VQA. 
We discussed with humans performing the task why they missed certain questions. The vast majority of errors made by humans were because they lacked sufficient external knowledge about certain characters or references to the image (e.g.~need to know that the Joker is a villain but Batman is a superhero) and were thus unable to figure out who or what was being referred to. 
To remedy this, we also granted humans access to the Internet and allowed them to search for references that they didn't recognize.
We observe that after granting Internet access, the external knowledge category of MMCOT jumped significantly. This shows that our questions are highly challenging and require external knowledge to answer. We note that performance for the Cause and Effect category seems high for all models when compared to other categories, but this is because it is a binary task where random performance is 50\%.

\begin{table}[t]
\newcolumntype{L}{>{\centering\arraybackslash}m{1.2cm}}
\captionsetup{skip=5pt}
\centering
\fontsize{8}{9}
\selectfont
\setlength{\tabcolsep}{3pt}
\renewcommand{\arraystretch}{1.3}
\resizebox{\textwidth}{!}{%
\begin{tabular}{l|LLLLLLLLLLLLLL} \hline \hline
Model &
  Total &Consistency (joint accuracy)&
  Solution-verified&Common objects&Relevant objects with unusual properties&
  Irrelevant objects with unusual properties&
  Distractors &
  Occlusion &
  OCR &
  Large number&
  Hallucination &
  External knowledge\\ \hline
Human &84.09&44.32&82.30&88.78&84.43&84.48&82.89&81.35&96.42&81.69&82.23&76.32\\\hline
LLaVA-Next-Llama3-8B     & 20.03 & 3.62  & 19.65 & 18.42 & 20.83 & 17.07 & 15.03 & 15.99 & 21.43 & 6.86  & 10.55 & 10.34 \\
LLaVA-Next-Qwen 110B     & 40.43 & 7.46  & 40.28 & 34.58 & 35.00 & 30.89 & 24.85 & 26.32 & 42.26 & 14.29 & 18.81 & 17.24 \\
VILA-Llama3-8B           & 8.66  & 1.71  & 8.31  & 10.69 & 9.38  & 8.13  & 8.28  & 8.91  & 5.36  & 4.57  & 11.47 & 3.45  \\
Mantis-8B                   & 5.25  & 1.07  & 4.67  & 5.54  & 5.21  & 3.25  & 4.60  & 4.05  & 5.26  & 1.14  & 7.80  & 3.45  \\
GPT-4V                   & 49.34 & 9.62  & 48.99 & 54.23 & 51.67 & 64.70 & 41.70 & 42.90 & 60.00 & 26.70 & \textbf{22.89} & 36.17 \\
GPT-4o + Captioning&\textbf{62.70}&\textbf{15.35}&\textbf{62.32}&\textbf{69.16}&\textbf{70.83}&\textbf{70.73}&\textbf{54.60}&59.72&63.10&41.14&19.27&\textbf{60.34}\\
GPT-4o& 62.18 & 12.15 & 61.97 & 68.90 & 66.45 & 64.70 & 53.90 & \textbf{59.90} & \textbf{71.86} & \textbf{44.88} & 13.30 & 58.62 \\
InternVL-Chat-V1.5-13B& 9.77  & 3.84  & 9.65 & 11.47 & 10.41 & 12.19 & 10.43 & 8.91  & 7.14  & 6.28  & 8.71  & 10.34 \\
Blip-2-FlanXXL-12B & 3.13  & 1.07  & 2.55  & 3.36  & 2.29  & 2.44  & 2.76  & 3.44  & 2.98  & 0.57  & 3.67  & 1.72  \\
InstructBLIP-Flan-T5-XXL-12B & 4.31  & 0.64  & 3.51  & 4.37  & 4.17  & 2.44  & 5.21  & 4.05  & 3.57  & 2.86  & 5.05  & 3.45  \\
MiniGPT4-Vicuna-13B & 3.73  & 0.21  & 3.27 & 3.12  & 4.58  & 0.00  & 5.52  & 3.04  & 4.76  & 1.14  & 12.84 & 3.45  \\
MiniGPT4-Llama2-7B          & 3.69  & 0.00  & 3.31  & 3.20  & 3.12  & 1.63  & 5.52  & 2.83  & 4.17  & 1.14  & 9.17  & 1.72  \\
mPLUG-Owl v2-9.2B             & 7.07  & 1.07  & 6.72  & 6.87  & 7.92  & 8.13  & 5.83  & 5.06  & 8.93  & 3.43  & 6.42  & 5.17  \\
mPLUG-Owl-7.2B                & 3.19  & 3.00  & 2.69 & 3.67  & 2.08  & 2.44  & 5.52  & 2.43  & 1.79  & 0.00  & 5.50  & 0.00  \\
CogVLM v2 (Llama3)-19B       & 8.73  & 0.21  & 8.23 & 9.44  & 9.17  & 7.31  & 7.97  & 6.07  & 8.92  & 4.00  & 11.00 & 0.00  \\
BEiT3-674M        &4.10&0.64&2.10&2.97&4.38&4.07&3.07&2.23&3.57&1.71&13.76&0.00\\
\hline \hline
 
\end{tabular}%
}
\caption{Zero-shot detailed result of MCOT across categories on JourneyBench dataset. GPT-4o+Captioning indicates using GPT-4o to solve MCOT using descriptive captions of the images also generated by GPT-4o.}
\label{tab:mcot_cate_results}
\end{table}
\subsection{Detailed MCOT Results Across Categories}
Table~\ref{tab:mcot_cate_results} presents detailed results of various SOTA vision-language models (VLMs) across different categories of our proposed JourneyBench MCOT dataset. Our JourneyBench MCOT dataset is divided into various categories to assess the performance of different state-of-the-art vision-language models (VLMs). For the MCOT task, GPT-4o achieves the highest performance across different aspects of the dataset, obtaining an overall accuracy of 62.18\%. It outperforms all other models in every category except for \textit{Hallucination} detection, where GPT-4V demonstrates the most promising performance.

GPT-4o's superior performance extends to the relevant objects with unusual properties (66.45\%) and irrelevant objects with unusual properties (64.70\%) categories, indicating its adeptness at managing complex and atypical visual information. Additionally, GPT-4o shows significant strength in the OCR category (71.86\%) and large numbers category (44.88\%). For the external knowledge category, GPT-4o achieves the highest score (58.62\%), demonstrating its proficiency in leveraging external information to enhance understanding and accuracy. Overall, GPT-4o stands out as the leading model in the MCOT task across the JourneyBench dataset, consistently outperforming other models in a wide range of categories. 
JourneyBench highlights GPT-4o's broad abilities to handle diverse and complex visual tasks across many different settings.

We also include the GPT-4o+Captioning result: first, we use GPT-4o to describe the image in detail, especially describing the number of each item in the image. Then, we input the question with the generated caption to GPT-4o together. However, this does not show a significant increase in the overall accuracy of the answers. The analysis in the table shows that the accuracy increased in all other categories except for the OCR and large number categories. This is possibly due to miscounting and misidentifying during the captioning phase.

\begin{table}[]
\newcolumntype{L}{>{\centering\arraybackslash}m{1.4cm}}
\centering
\captionsetup{skip=5pt}
\fontsize{8}{9}
\selectfont
\setlength{\tabcolsep}{3pt}
\resizebox{0.8\textwidth}{!}{%
\begin{tabular}{l|LLLLLLL}
\hline \hline
\multicolumn{8}{|c|}{Text Retrieval} \\ \hline 
Categories &  ALBEF-210M &BEiT3-674M&CLIP-430M&X2\_VLM-590M&InternVL-C-13B&InternVL-G-14B&
OpenCLIP-CoCa-13B\\ \hline
Incorrect physics rules &  \textbf{69.57} &  67.39 &  60.87 &  58.70 &  67.53 &  68.08 &68.22 \\
Incorrect biological rules & 63.38 &  \textbf{67.61} &  54.93 &  53.52 &  63.53 &  64.05 &64.11 \\
Misplacement &  71.60 &  69.14 &69.14 &  61.73 &  71.66 &  \textbf{71.81} &  70,14 \\
Strange animal &  \textbf{73.15} &  73.15 &  63.89 &  61.11 &  70.96 &  71.84 &  74.11 \\
Unexpected behavior &  76.47 &  77.65 &  72.55 &  70.98 &  75.20 &  \textbf{78.31} &  70.84 \\
Unusual food &  68.75 &  70.83 &  72.92 &  70.83 &  72.02 &  \textbf{74.69} &  71.92 \\
Strange indoor objects &  60.00 &  58.46 &  56.92 &  58.46 &  62.33 &  62.94 &\textbf{63.77} \\
Strange scene &  60.00 &56.00 &  49.60 &48.00 &56.38 &57.89 &\textbf{61.00} \\
Unusual construction & 51.52 &\textbf{53.03} &43.94 &45.45 &52.15 &52.87 &52.39 \\
Incorrect usage & 60.00 &  \textbf{74.29} &  71.43 &  60.00 &  70.97 &  70.49 &  72.14 \\
Unusual attributes or accessories &  60.70 &  60.95 &  58.46 &  58.21 &  63.28 &  \textbf{63.59} &  55.74 \\ \hline
\multicolumn{8}{|c|}{Image Retrieval}\\ \hline 
Categories & ALBEF-210M&BEiT3-674M&CLIP-430M& X2\_VLM-590M&InternVL-C-13B&InternVL-G-14B&OpenCLIP-CoCa-13B \\ \hline
incorrect physics rules &  48.70 &  52.17 &  53.91 &  57.39 &  59.70 &  \textbf{60.52} &  53.20 \\
incorrect biological rules &  46.20 &  55.21 &  50.70 &  59.72 &  59.78 &  \textbf{60.48} &  53.02 \\
misplacement &  56.79 &  65.43 &  57.78 &  59.26 &  66.72 &  \textbf{67.44} &  53.18 \\
strange animal &  55.56 &  60.74 &  47.04 &  66.11 &  63.60 &  \textbf{64.21} &  60.47 \\
unexpected behavior &  63.37 &  67.37 &  64.16 &  68.63 &  72.28 &  \textbf{72.81} &  61.14 \\
unusual food &  60.83 &  73.33 &  70.83 &  74.17 &  76.23 &  \textbf{76.97} &  59.88 \\
strange indoor objects &  49.54 &  54.77 &  47.38 &  52.92 &  57.50 &  \textbf{58.00} &  49.61 \\
strange scene &  42.64 &  47.52 &  40.08 &  47.52 &  50.76 &  \textbf{51.81} &  43.26 \\
unusual construction &  27.58 &  38.18 &  36.97 &  36.67 &  41.55 &  \textbf{42.36} &  28.08 \\
incorrect usage &  64.57 &  74.29 &  71.43 &  68.57 &  76.50 &  \textbf{77.12} &  62.84 \\
unusual attributes or accessories &  48.51 &  52.89 &  49.65 &  53.98 &  57.78 &  \textbf{59.28} &  49.13 \\ \hline \hline
\end{tabular}%
}

\caption{Zero-shot detailed results (R@1) of Retrieval across categories on our proposed JourneyBench dataset.}
\label{tab:CR_cate_results}
\end{table}
\subsection{Detailed Retrieval Results Across Categories}
We present the performance of different state-of-the-art (SOTA) retrieval models on our proposed JourneyBench retrieval dataset in Table~\ref{tab:CR_cate_results}. The dataset is annotated into 11 categories, ranging from ``incorrect physics rules'' to ``unusual attributes or accessories,'' to challenge the retrieval models' 
performance. 

Overall, for the text retrieval task, InternVL-14B and OpenCLIP-CoCa generally demonstrate strong performance across most categories. In the ``incorrect usage'' category, BEiT3 obtains the highest R@1 score of 74.29\%, which is slightly higher than InternVL-14B (70.49\%) and OpenCLIP-CoCa (72.14\%).

For the image retrieval task, InternVL-14B outperforms all the models across all categories of our proposed JourneyBench dataset. Across both retrieval tasks, InternVL-14B frequently appears as one of the top performers in handling diverse and complex categories within the JourneyBench dataset.

\begin{table}[]
\newcolumntype{L}{>{\centering\arraybackslash}m{1.2cm}}
\captionsetup{skip=5pt}
\centering
\fontsize{8}{9}
\selectfont
\setlength{\tabcolsep}{3pt}
\resizebox{\textwidth}{!}{%
\begin{tabular}{l|LLLLLLLLLLLL} \hline \hline
Model &Overall&
  Incorrect physics rules&
 Incorrect biological rules&
  Misplacement &
  Strange animal&
  Unexpected behavior&
  Unusual food&
  Strange indoor objects&
  Strange scene&
  Unusual construction&
  Incorrect usage&
  Unusual attributes or accessories\\ \hline
Human&85.71&83.57&84.80&89.94&86.47&93.13&97.98&83.21&84.23&80.15&93.85&84.01\\\hline
OpenCLIP-CoCa (Vit-L)-13B    &21.59& 22.32 & 19.25 & 28.24 & 19.43 & 26.87 & 29.65 & 23.16 & 17.89 & 20.71 & 22.40 & 21.93 \\
LLaVA-Next-Llama3-8B     &28.69& 33.90 & 28.98 & 36.90 & 28.21 & 32.63 & 35.95 & 27.21 & 24.93 & 21.65 & 32.09 & 29.02 \\
LLaVA-Next-Qwen110B&27.18&35.57&22.96&33.94&25.03&32.60&33.62&25.98&21.19&24.51&23.47&28.461 \\
GPT-4o&32.56&37.05&\textbf{35.63}&\textbf{48.29}&32.98&41.97&38.52&\textbf{35.82}&25.76&20.18&\textbf{51.32}&29.06 \\
GPT-4V&11.24& 12.44&17.29&19.43&10.34&17.57&10.33&8.99&7.67&6.81&22.34&9.84\\
InstructBLIP-Flan-T5-XXL-12B &26.00& 0.74&0.53&0.10&1.56&0.03&0.10&0.02&0.81&0.02&0.02&0.2512 \\
MiniGPT4-Llama2-7B&20.91&27.82&23.65&24.75&23.50&25.06&21.32&22.27&16.06&15.77&29.71&18.96 \\
MiniGPT4-Vicuna-13B&16.21&20.13&18.18&21.85&15.28&20.48&20.57&20.87&13.32&11.98&19.42&16.09\\
mPLUG-Owl v2-9.2B&26.74&27.39&28.03&37.34&31.62&38.46&24.33&25.46&21.24&17.77&33.80&24.68\\
mPLUG-Owl-7.2B&14.68&18.04&10.25&21.52&16.46&18.96&13.74&17.25&13.20&12.88&15.06&13.77\\
CogVLM v2 (Llama3)-19B&30.31&33.39&33.01&39.15&28.41&41.34&\textbf{41.06}&30.72&20.81&22.92&45.13&28.84\\ 
BEiT3-674M	&30.90&33.25 &27.08&45.27&27.12&38.39&31.59&28.49&24.64&\textbf{27.24}&28.45&31.78\\

Mantis\_Idefics2-8B &33.34&32.88&37.47&43.91&\textbf{34.57}&\textbf{42.42}&34.41&35.46&26.67&25.18&35.62&31.09\\

VILA-8B&\textbf{33.79}&\textbf{39.05}&34.02&43.92&32.32&39.45&38.42&32.23&\textbf{28.12}&23.02&37.59&\textbf{34.61}\\
\hline \hline
\end{tabular}%
}
\caption{Zero-shot detailed results (CIDEr scores) of imaginary image captioning on our proposed JourneyBench dataset. The human performance is computed by holding out one of the five annotated captions as prediction and computing the score using the rest as ground truth.}
\label{tab:ugc_cate_results}
\end{table}

\subsection{Detailed Captioning Results Across Categories}
Table~\ref{tab:ugc_cate_results} presents the zero-shot detailed results (CIDEr scores) of various models on the imaginary image caption generation task on our proposed JourneyBench dataset. The table evaluates the models across eleven categories: incorrect physics rules, incorrect biological rules, misplacement, strange animal, unexpected behavior, unusual food, strange indoor objects, strange scene, unusual construction, incorrect usage, and unusual attributes or accessories. 

To set up the benchmark performance and to illustrate the challenging nature of our proposed dataset, we also assess human performance on imaginary image caption generation. We consider this an upper bound on the captioning performance. To compute our human upper bound, we consider the set of captions for each sample. We treat each ground truth caption as a machine generated caption and use the remaining ground truth captions to compute the CIDEr score for the ground truth caption. We repeat this for every ground truth caption in each set. We find that the human CIDEr score is far higher than any machine captioning approach. This indicates to us that our captioning task is sensible (i.e.~humans agree with one another on the task), but very challenging for machines given the performances shown. Human written captions achieve the highest scores in all categories of the dataset. Following human, GPT-4o, VILA and Mantis\_Idefics2 models show strong performances. GPT-4o outperforms other models in misplacement (48.29\%), strange indoor objects (35.82\%) and incorrect usage (51.32\%). VILA achieves highest scores among the models in incorrect physics rules (39.05\%), strange scene (28.12\%) and unusual attributes or accessories (34.61\%). Mantis\_idefics2 obtains highest scores in incorrect biological rules (37.47), strange animal (34.57\%) and unexpected behavior (42.42\%). 
However, CogVLM v2 (Llama3) outperforms all the models in unusual food category.  Our results highlight the varying capabilities of different models in generating captions for unusual and complex scenarios within the JourneyBench dataset. While GPT-4o, VILA and Mantis\_Idefics2 emerge as strong performers across multiple categories, the human upper bound indicates there is significant room for improvement in achieving human-like caption generation on imaginary generated images. One possible reason for this low performance is that models rely too heavily on their language biases for captioning which prevents them from describing objects or actions that are unusual.

\section{Annotation}

\subsection{Annotation Details}

\begin{figure}[ht!]
    \centering
    \captionsetup{skip=5pt}
    \resizebox{1\textwidth}{!}{
    \includegraphics{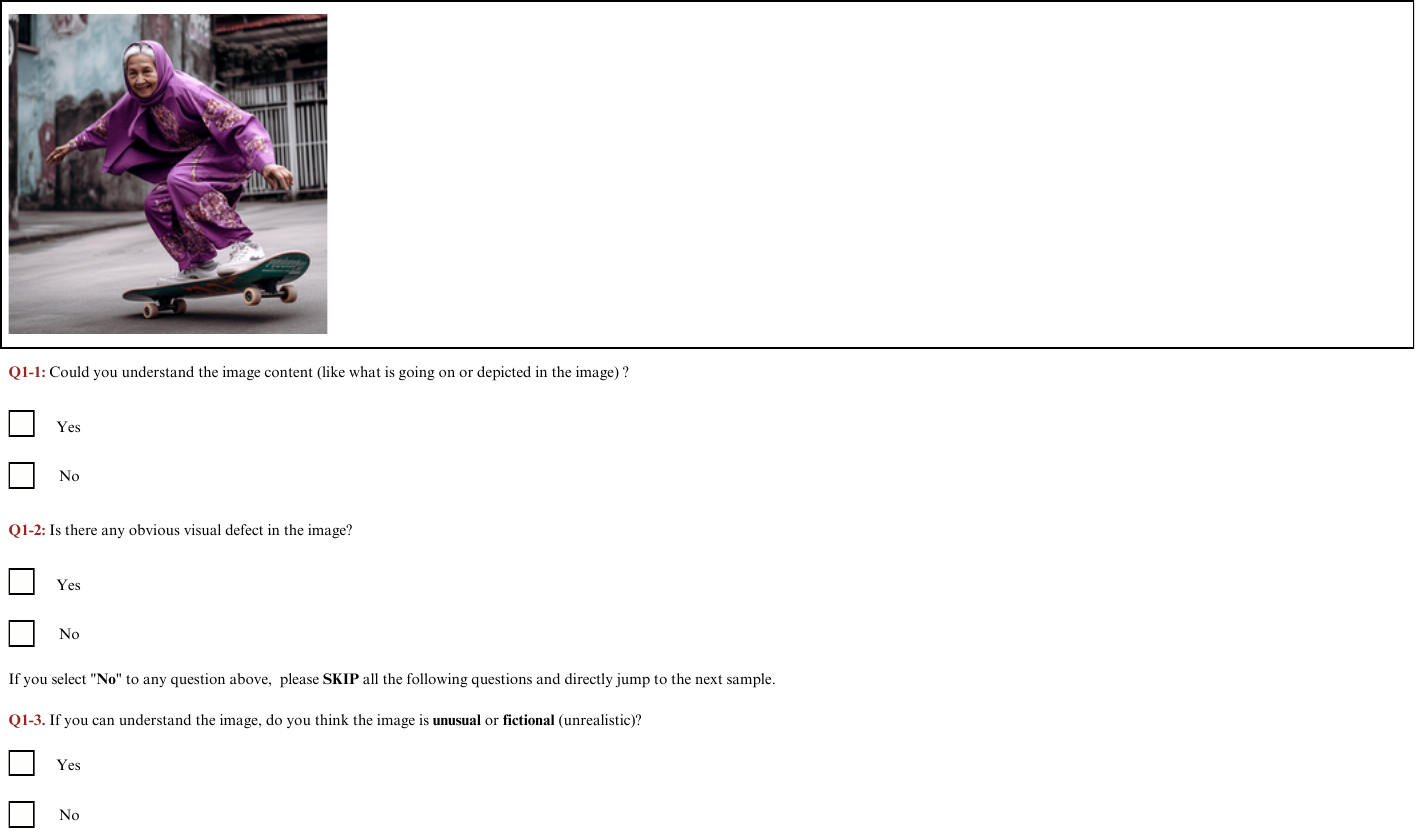}
    }
  \caption{Annotation interface for imaginary image filtering.}
  \label{fig:imaginary-image-ui}
\end{figure}

\subsubsection{Image Filtering}

After retrieving images, human annotators filter the image set harvested using our retrieval process based on three key criteria: the images must be \textbf{unusual} or \textbf{fictional} (unrealistic), and they must also be \textbf{comprehensible}. 
Unusual images depict scenarios outside of everyday experiences, feature unexpected juxtapositions of objects, or include visually striking elements. 
Fictional images, on the other hand, present unrealistic or impossible scenes in the real world (\textit{e.g.~}an elephant standing on macaroons). 
However, we also enforce that the images are free of artifacts and understandable to humans to describe. This ensures a balance between creating challenging scenarios and maintaining the ability to reliably identify specific weaknesses in model reasoning or understanding. As shown in Figure \ref{fig:imaginary-image-ui}, we assess this by directly asking annotators a set of questions, including ``Can you understand the content in the image?", ``Is there any obvious visual defect in the image?'' and ``If you can understand the image, do you think the image is unusual or fictional (unrealistic)?''.
We understand identifying imaginary images may be subjective, so for every image, we hire at least four Amazon Mechanical Turk (MTurk) crowd-sourced annotators to answer those questions to determine, and the 4/4 agreement is achieved in more than 72$\%$ of the cases. 

\begin{figure}[ht!]
    \centering
    \captionsetup{skip=5pt}
    \resizebox{\textwidth}{!}{
    \includegraphics{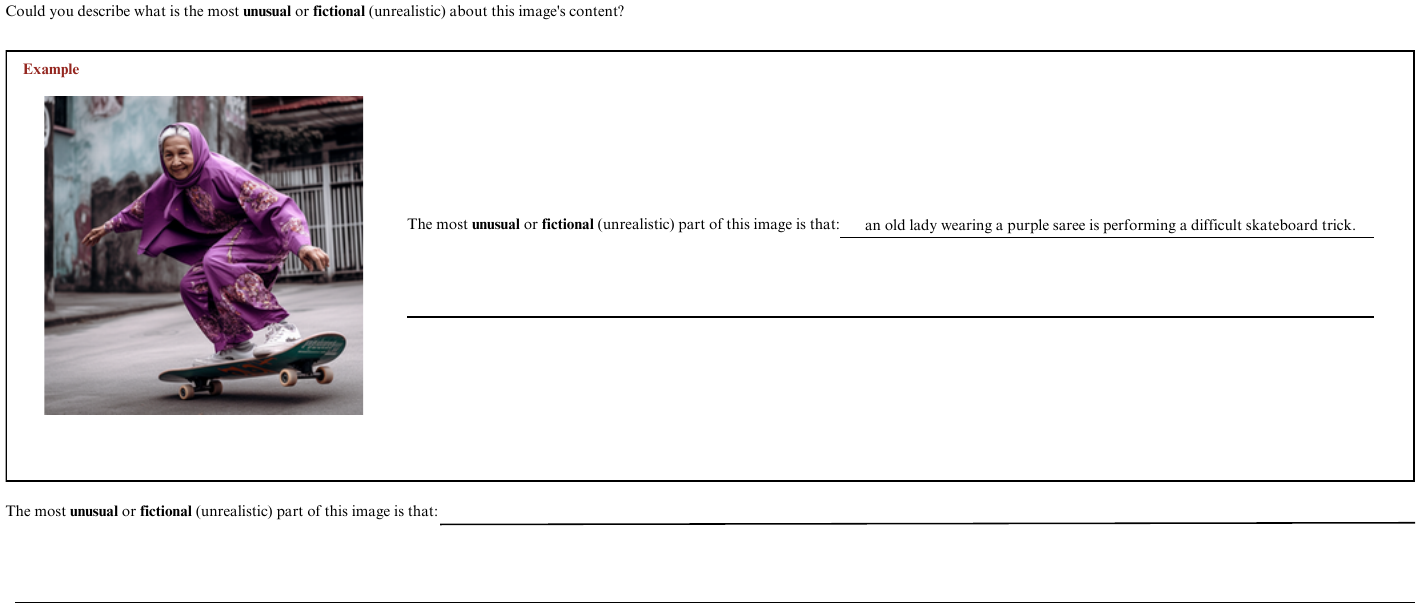}
    }
  \caption{User interface for imaginary image captioning.}
  \label{fig:imaginary-image-caption-ui}
\end{figure}

\subsubsection{Image Captioning}

In JourneyBench, we also include a captioning task, but seek to test the models' abilities to understand and caption imaginary images. For this task, we require models to generate a single-sentence description of an image highlighting elements that make it imaginary. 
We first want to obtain the ground-truth image captions. Hence, for each collected imaginary image, we ask eight MTurk annotators to describe the most unusual or fictional part of the image in one sentence, as in Figure \ref{fig:imaginary-image-caption-ui}. To avoid subjective biases among annotators, those generated descriptions are further verified by another group of four experienced MTurk annotators to vote to determine whether they agree with the description. For every image, we only reserve the top five highest-voted descriptions, and each one must obtain at least two votes from the verifiers.
If an image does not have five descriptions, each with at least two votes, then we believe there may not be enough agreement to determine the description, and the image is discarded.

\subsection{Quality Assurance} 

For every step requiring annotations during our data collection process of JourneyBench, we prepare detailed instruction manuals with many examples.
Given the challenging nature of our tasks, for each annotation step, we also hire at least two master annotators to supervise the annotation results for each batch to quickly verify the results by poor annotators.
Defective annotations are sent back for re-correction with instructions, and annotators with quality annotation history are assigned more batches of data for annotation. 
Collectively, our annotators spent more than $2,200$ hours annotating JourneyBench. 
To help identify easy or low-quality samples, we have annotators verify the data quality of every annotated sample. 
To avoid human biases, we also apply adversarial models for every sample across five tasks. For instance, for MCOT questions, we leverage LLMs to guess answers and remove samples where language-only models can guess the ground-truth answers.


\subsubsection{Adversarial Filtering}

\textbf{Filteirng via VLMs and LLMs: }In order to ensure the challengeness and quality of our VLU tasks like VQA (HaloQuest), MCOT and Multi-image VQA, we inference a spectrum of VLMs of various sizes to those tasks. We filter to samples where most of VLMs can easily obtain the correct answer with high confidence scores and regard those samples as ``too easy" and modify them to be more challenging or directly remove them. Additionally, to further ensure there is not shallow bias or shortcut in our data, we also apply language-only models to inference over these tasks and move the ones language-only models can score correctly. 

\begin{figure}
    \centering
\includegraphics[width=\textwidth, keepaspectratio]{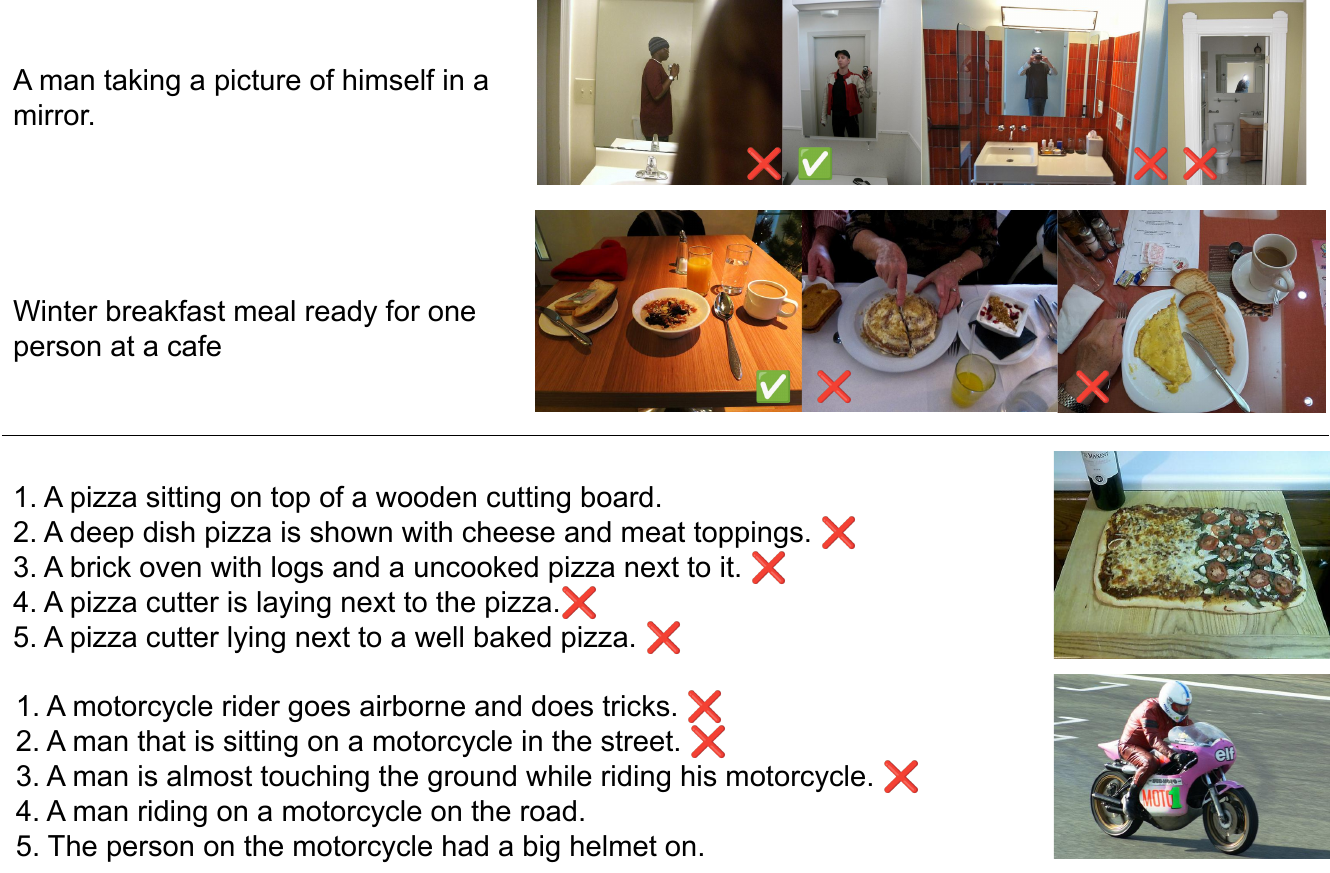}
  \caption{\textbf{Top figure: false negatives in MS COCO 5K image retrieval.} These images from different data points fit the description of the same text. They are indistinguishable from the ground truth image (labeled by the green checkmark) even from the human perspective. \textbf{Bottom figure: false positives in ECCV Caption image retrieval.} A significant number of texts matched to the image by the annotation describe scenes similar to but different from the ground truth image (the red cross mark labels these captions). Evaluation results on these data points will be inaccurate.}
  \label{fig:mscoco_fn}
\end{figure}

\textbf{Filtering False Positives/Negatives: }
Current datasets commonly used in the field often grapple with issues such as inconsistencies, false negatives, ambiguities, and more. As an illustration, Figure \ref{fig:mscoco_fn} highlights examples of false negatives within the widely-used MS COCO 5K image retrieval dataset \cite{mscoco}, a problem largely stemming from the sampling process from the original captioning dataset. Although there have been efforts to rectify these inaccuracies \cite{eccv-caption} they have inadvertently introduced false positives, which were non-existent in the original dataset. Such examples are also depicted in Figure \ref{fig:mscoco_fn}. 

In contrast, our retrieval dataset, despite also being sampled from our captioning dataset, primarily utilizes generated images that inherently minimize the occurrence of false negatives due to the highly randomized combination of elements within these images, a point we discussed thoroughly in the main paper. For example, in the second instance from Figure \ref{fig:mscoco_fn}, the conventional dataset images involve highly related elements like "food" and "table", with a high frequency of appearing in other data points, too. In our dataset, rare combinations such as "cat" and "kimono", "CPU" and "soup", or "sander" and "donuts" (more detailed analysis in Section \ref{sec:triplet}), demonstrate a broader and more varied semantic range, with a much lower chance of having overlapping topic words among images. Finally, the prompt-based generated images on MidJourney \cite{midjourney} always have prompts available, which are accurate descriptors of the images, allowing us to group images by prompt to easily verify and filter false negative image-text pairs for retrieval tasks. Consequently, the likelihood of semantically similar images existing in our retrieval dataset is significantly reduced, minimizing the risk of false negatives.

\subsubsection{Machine Focus v.s. Human focus}
\begin{figure}
    \centering
\includegraphics[width=\textwidth, keepaspectratio]{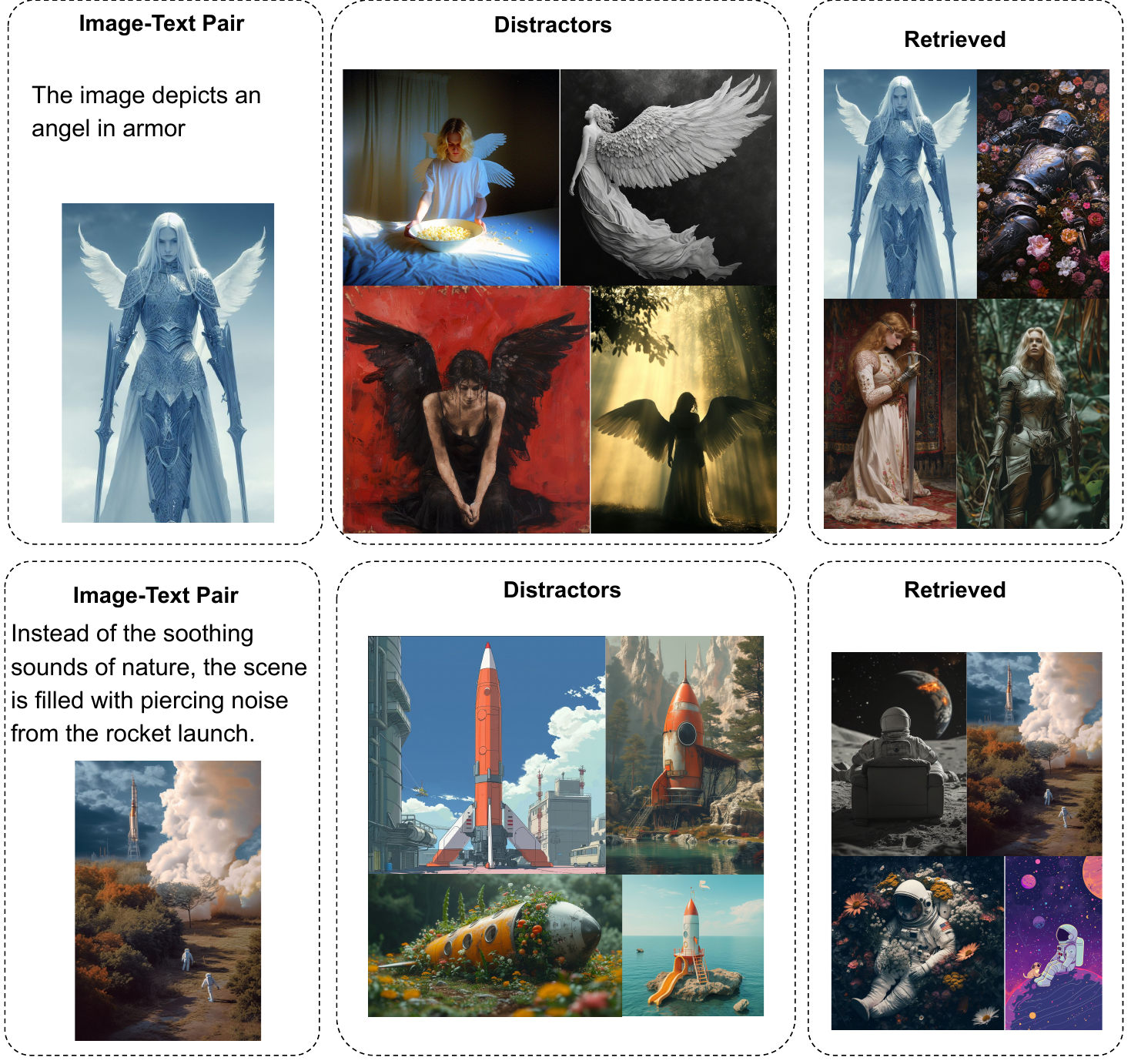}
  \caption{\textbf{Comparison between machine and human focus of images.} The distractors are collected by annotators to be semantically similar to the image. However, models sometimes do not retrieve these distractors because they focus on different aspects of the text.}
  \label{fig:machine_focus}
\end{figure}
\begin{figure}
    \centering
\includegraphics[width=\textwidth, keepaspectratio]{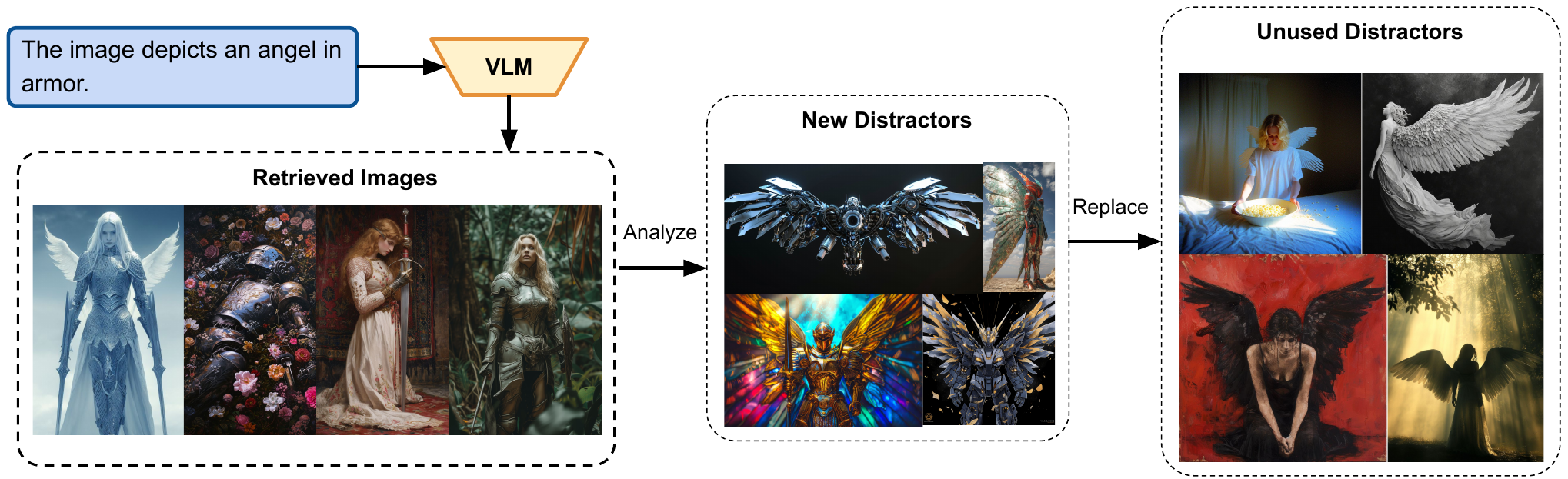}
  \caption{\textbf{One round of adversarial annotation.} The annotators analyze the retrieved images by the VL models, then collect new distractors that are closer to the models' judgment to replace the unused ones.}
  \label{fig:improve_distractor}
\end{figure}
A large semantic domain for images, despite minimizing false negatives in the annotation, comes at a cost of lower retrieval difficulty, since all images/texts are highly distinct. To address this, we introduced sample-specific distractors in our retrieval dataset, as detailed in the main paper. These distractors, collected by human annotators, are both visually and semantically similar to the target images, differing only subtly to challenge the retrieval models without being misclassified as true positives.

However, the decision-making process of VL models does not always align with human judgment, as illustrated in Figure \ref{fig:machine_focus}. The distractors collected by humans focus on certain elements like "angle" and "rocket", VL models might retrieve based on other features such as "armor" and "nature". To maintain a high level of retrieval difficulty, it is crucial to consider the perspective of VL models.

To bridge the gap between human and machine perception, we implement a multi-stage annotation process. Initially, we designate two sets of VL models — the "signal" set and the "test" set. We first evaluate the signal set models using the image retrieval dataset that includes the distractors. A distractor is deemed ineffective if none of the models retrieve it among the top five results. These ineffective distractors are then replaced based on an analysis of the top images retrieved by the models. Subsequently, we test the models on the datasets both before and after these adjustments to demonstrate the changes’ effectiveness. This approach harmonizes the focal points of both humans and machines in assessing the images. Practically, we conduct two rounds of this improvement process, selecting two models each for the signal and test sets, while the remaining models are excluded from the annotation process.

\section{Dataset Statistics}
\subsection{General Statistics}

Overall, JourneyBench has $13,631$ unique image-text samples across five tasks, which consist of $12,405$ unique images and $13,664$ unique text. JourneyBench includes 2,600 image-question pairs for complementary multimodal chain-of-thought, categorized into 10 fine-grained types based on visual contexts and multimodal co-referencing. All collected images in JourneyBench fall into 11 fine-grained categories based on their level of unusualness or fictionality.
For multi-image VQA, there are 316 image-question pairs across three fine-grained categories. 
We note that this is larger than the recent multi-image VQA evaluation benchmark (217 samples) in Mantis \cite{mantis}.
The image captioning dataset contains 1,000 images paired with 5,000 captions, with each image having five captions. 
For visual question answering, JourneyBench comprises 7,748 image questions, categorized into three fine-grained types of hallucination triggers. 
The fine-grained cross-modal retrieval task contains two subtasks. For image-to-text retrieval, there are 1,000 query images paired with 11,121 texts, averaging five positive texts (ground-truth captions) and six negative texts (sample-specific text distractors) per image. For text-to-image retrieval, there are 1,000 samples, each with five ground-truth captions, resulting in approximately 5,000 query texts against 6,323 images. Each sample has one ground-truth matching image and five negative images (sample-specific image distractors). 
\subsection{Categories Analysis}
\begin{figure}
    \centering
\includegraphics[width=\textwidth, keepaspectratio]{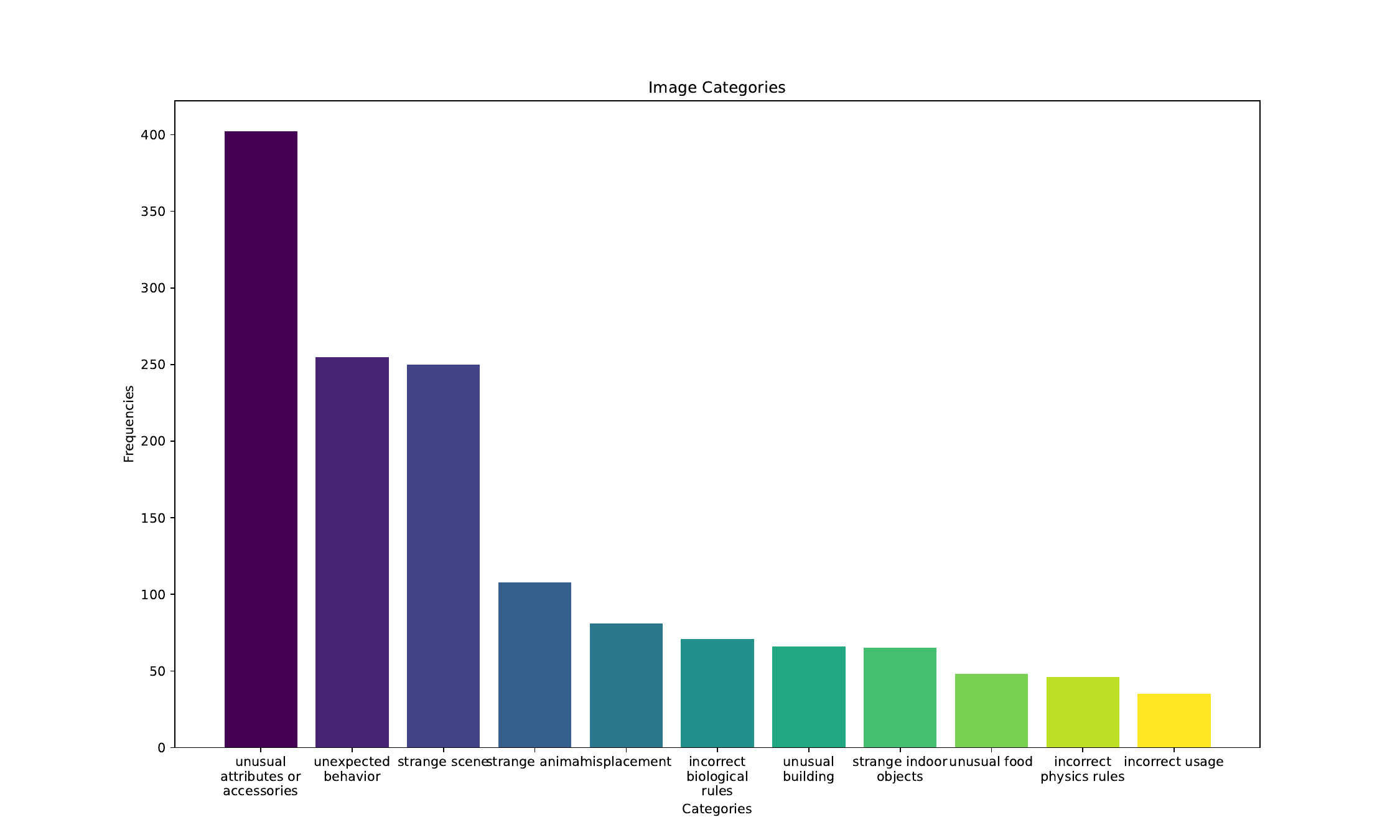}
  \caption{Frequency of categories in Imaginary Image Captioning. The categories describe the unusualness of the images.}
  \label{fig:uct_catepgories_stats}
  \end{figure}
\textbf{Imaginary Image Categories.}
Our imaginary image captioning dataset comprises a variety of imaginary images, classified using a set of unique categories for analysis purposes. Figure \ref{fig:uct_catepgories_stats} displays the frequency of each category. We manually annotate each image with up to two of the 11 available categories. The diversity of scenarios challenges the models to thoroughly understand each image in order to perform effectively. Detailed examples for each category are provided in the qualitative examples section, illustrating the breadth of unusual cases that test the models' interpretive abilities.

\begin{figure}
    \centering
\includegraphics[width=\textwidth, keepaspectratio]{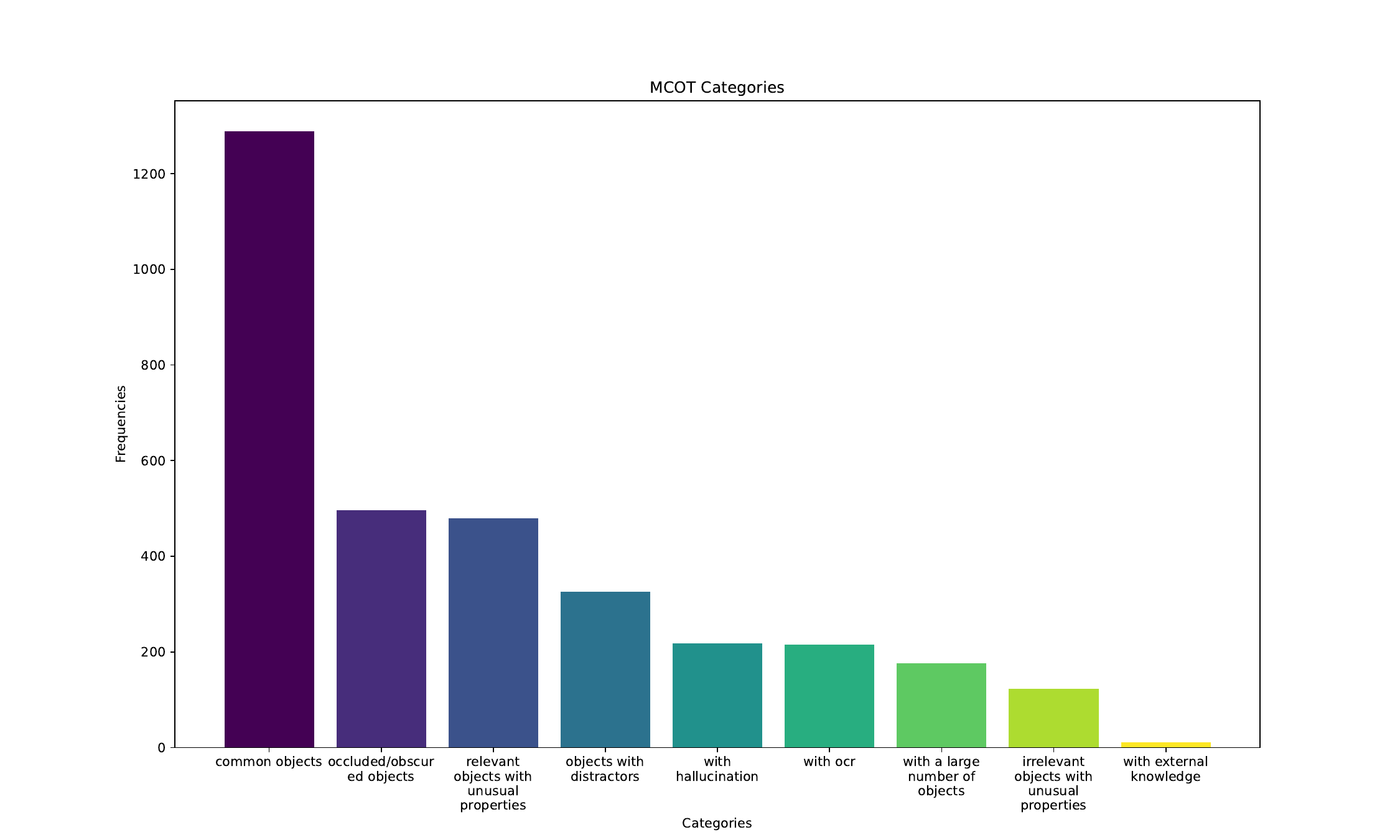}
  \caption{Frequency of categories in MCOT.}
  \label{fig:category_stats}
\end{figure}

\textbf{MCOT Co-referencing Categories.}As detailed in the main paper, our MCOT dataset necessitates that models reference the accompanying images to solve the math word problems presented. The questions are designed in various ways to reference images, creating diverse testing scenarios. Each data point is manually categorized to analyze the relationship between the questions and images. Figure \ref{fig:category_stats} illustrates the distribution of these categories within the MCOT dataset. Additional examples from each category are available in the qualitative examples section, showcasing the range of co-referencing strategies employed in the dataset.

\begin{figure}
    \centering
    
    \scalebox{0.6}{
\includegraphics[width=\textwidth, keepaspectratio]{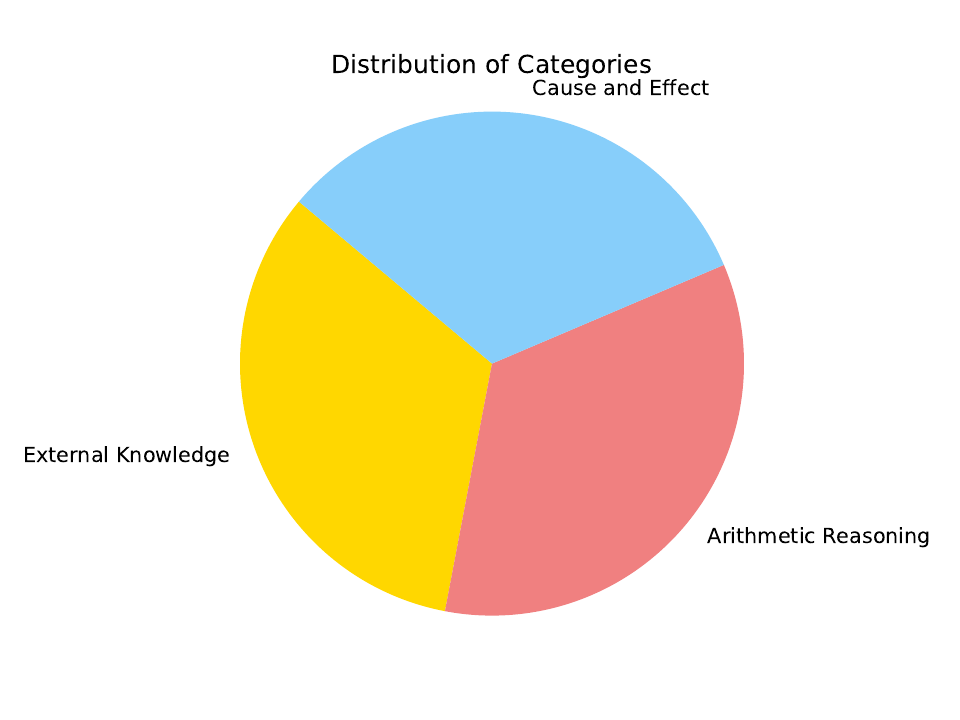}
}
  \caption{Frequency of categories in Multi-Image VQA.}
  \label{fig:multiimage_stats}
\end{figure}

\textbf{Multi-image Categories}
Our multi-image VQA dataset contains 2 tasks: multi-image MCOT and cause and effect, with multi-image MCOT further divided into two subcategories: arithmetic
reasoning and external knowledge. In Figure \ref{fig:multiimage_stats} we show the percentage of each category in the dataset.

\begin{figure}
    \centering
\includegraphics[width=\textwidth, keepaspectratio]{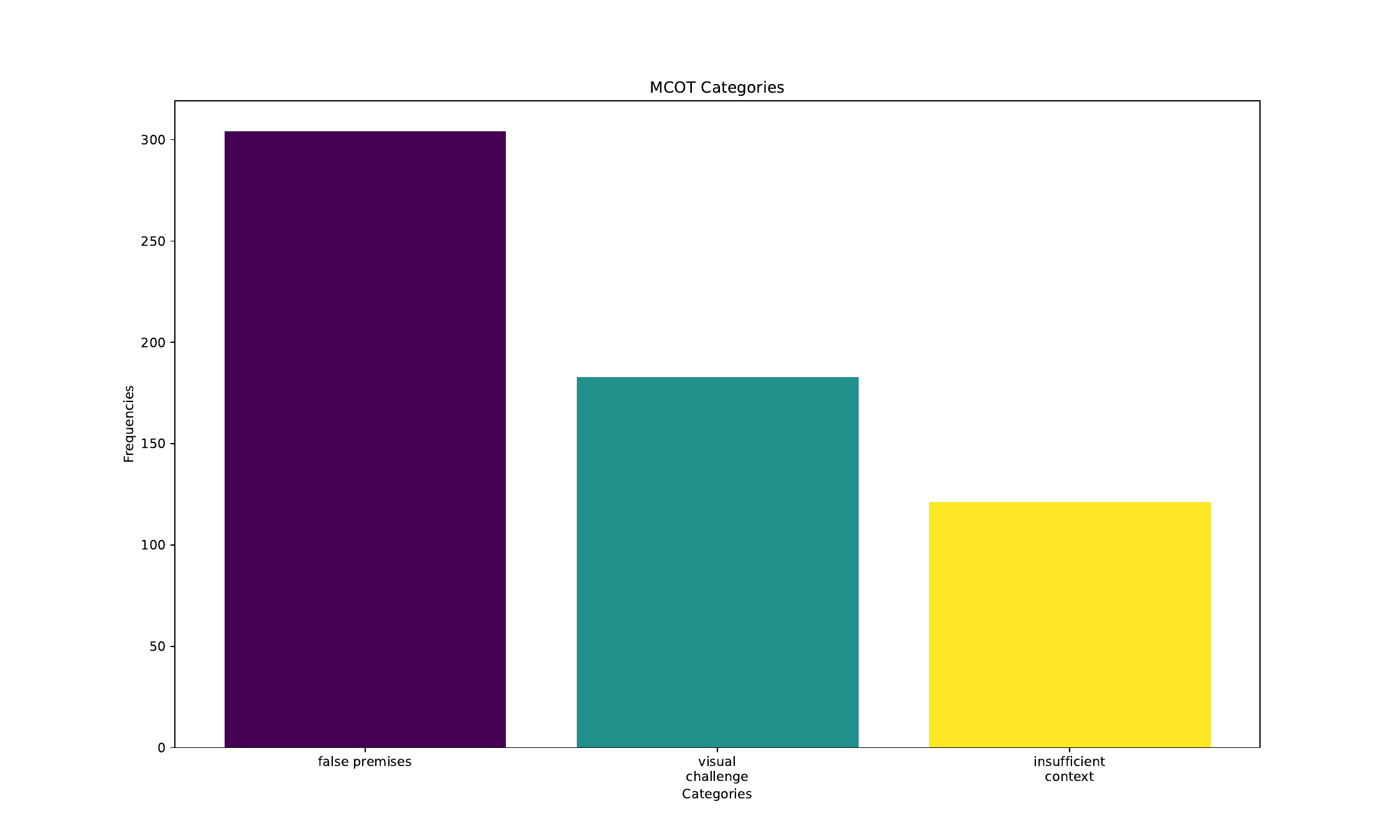}
  \caption{Frequency of categories in HaloQuest.}
  \label{fig:haloquest_categories}
\end{figure}

\textbf{HaloQuest Categories}
Similar to other tasks, each HaloQuest data point is associated with a hallucination category describing the type of challenging scenario the question is testing. We show the distribution in Figure \ref{fig:haloquest_categories}.
\section{Unusual Visual Scenes}

\subsection{Unusual Triplet Analysis}
\label{sec:triplet}
To illustrate the unusualness of JourneyBench images, we directly compared them with existing benchmarks such as MS-COCO. We randomly sampled 100 images from JourneyBench and other benchmarks, then had experienced annotators manually extract visual triplets contributing to the images' composition. These triplets, similar to unit triplets in conventional visual scene graphs, represent the visual makeup of the images. Our goal was to quantify the unusualness of these images by assessing the unusualness of the triplets based on common sense knowledge.

\begin{table}[]

\centering
\resizebox{0.5\textwidth}{!}{%
\begin{tabular}{l|c|c} \hline \hline
Dataset& Human Verify & ConceptNet \\ \hline
JourneyBench& 8.00	& 6.00 \\
COCO& 72.00	 & 68.00 \\ \hline \hline
\end{tabular}
}
\caption{{Related triplets in images.} We extract triplets of subjects from images and verify their relation through human annotators and ConceptNet. JourneyBench has significantly fewer related triplets in images, indicating the unusualness of the images.}
\label{tab:triplets}

\end{table}
To evaluate the unusualness of these triplets, we used two methods. First, another group of three experienced annotators examined the triplets and voted on whether each was unusual. The label for each triplet was determined by the highest-voted option. To minimize human bias, we also employed a second approach using ConceptNet \cite{liu2004conceptnet}\footnote{www.github.com/ldtoolkit/conceptnet-lite}, an external knowledge graph database. We queried ConceptNet to check if each extracted triplet existed within its database. This involved projecting the subject and object of each triplet into ConceptNet and verifying if a relationship aligned with our extracted triplet. 
As shown in Table \ref{tab:triplets}, the majority of the triplets extracted from JourneyBench images were deemed unusual by both evaluation methods. This confirms the distinctiveness and significance of the image distribution in JourneyBench.

\subsection{Language Prior Analysis}

As mentioned previously, existing benchmarks consist of everyday images, which are often utilized for models' training and evaluation. This may cause existing models to develop biases of common visual compositions. Therefore, in reasoning, existing models may not fully examine the visual input information but can still resolve the task correctly based on prior knowledge. However, in edge cases in the real world, this would lead to serious application mistakes and consequences. To investigate this issue further, we directly apply language-only models to JourneyBench tasks and existing popular datasets for comparison.

\begin{table}[]
\centering
\resizebox{0.5\textwidth}{!}{%
\centering

\begin{tabular}{@{}lcccl@{}}
 \hline \hline
                       & \multicolumn{2}{c}{MCOT}                                              & \multicolumn{2}{c}{VQA}                \\ \midrule
Model                  & \multicolumn{1}{l}{JourneyBench-MCOT} & \multicolumn{1}{l}{ScienceQA} & \multicolumn{1}{l}{HaloQuest} & VQA v2 \\ \midrule
GPT-4o                 & 62.18                                 & 91.04                         & 68.10                         & 81.84  \\
GPT-4o (Language-only) & 16.64                                 & 83.90                         & 20.82                         & 61.28  \\ 
 \hline \hline
\end{tabular}
}
\caption{Comparing the effect of removing the visual elements from datasets. MCOT and HaloQuest show a significant performance drop, indicating the strict complementing relationship between our dataset's visual and textual elements.}
\label{tab:language-only-performance}
\end{table}

\subsubsection{LLM performance on MCOT versus ScienceQA}
For comparison, we infer a language-only GPT-4o, over the JourneyBench MCOT dataset and another existing MCOT dataset, ScienceQA~\cite{scienceqa}. From Table \ref{tab:language-only-performance}, we can observe that language-only GPT-4o can only score 16.64$\%$ on our MCOT dataset but can achieve up to 83.9$\%$ on ScienceQA.

\subsubsection{LLM performance on HaloQuest versus VQA v2} 

We further compare language-only GPT-4o over HaloQuest versus VQA v2, a popular VQA task. From Table \ref{tab:language-only-performance}, we can observe that language-only GPT-4o can achieve much lower performance on HaloQuest compared with VQA v2. 

Most importantly, the performance drop between GPT-4o and GPT-4o (language-only) is much larger on JourneyBecnh and much smaller on existing datasets. It is problematic that without critical visual input information, language-only models can still achieve high performances. This indicates that the underlying visual composition aligns with the models' prior knowledge or biases; thus, the visual information becomes redundant or trivial.

\begin{figure}[ht!]
    \centering
    \captionsetup{skip=5pt}
    \scalebox{0.4}{
    \includegraphics{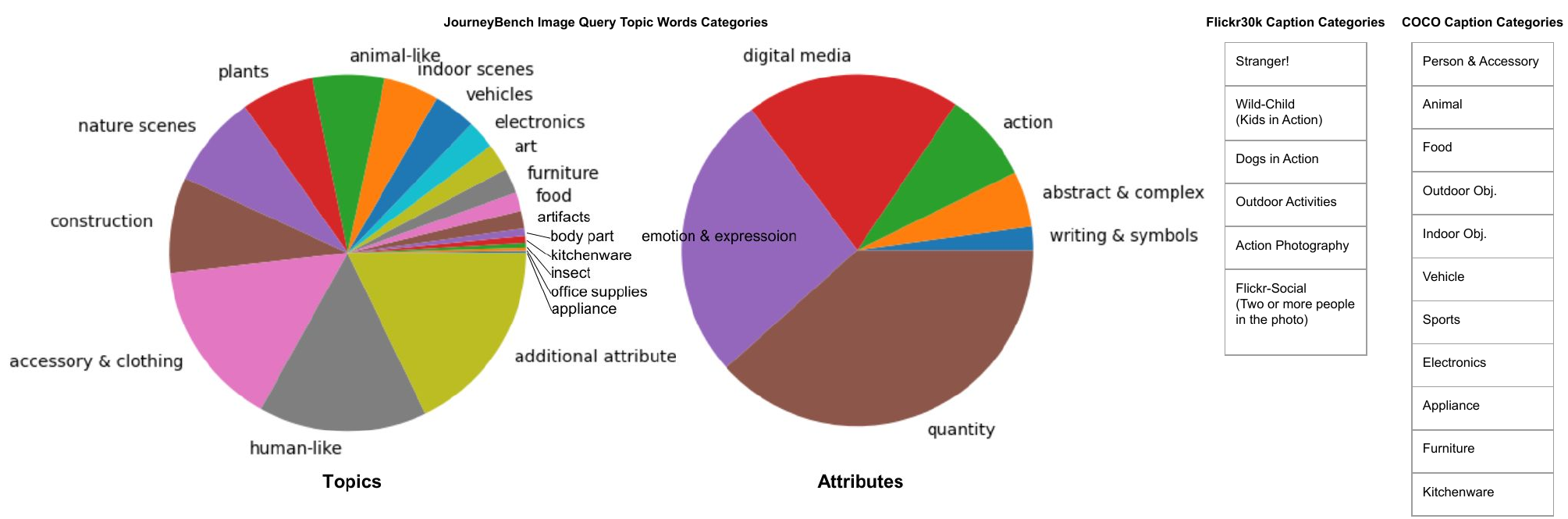}
    }
  \caption{Topics and attributes of JourneyBench data comparing to Flickr30k and COCO Caption. Our dataset covers a much wider range of topics.}
  \label{fig:image-topic-words}
\end{figure}
\section{Image Diversity Analysis}
\subsection{Image Topic Words Comparison}

We aim to create a VLU benchmark featuring challenging and diverse imaginary images, including unusual, abstract, and complex ones, by leveraging the advantages of prompt-based generated images. Initially, we followed the approach outlined in \cite{whoops} to handcraft prompts for generating images. However, we encountered difficulties avoiding a biased image distribution and ensuring high image quality. Instead, we discovered that utilizing metadata to \textit{retrieve} prompt-based generated images from a larger crowd-based platform provided higher quality and a more diverse distribution of images. Thus, we developed web scraping tools to analyze metadata from Midjourney\footnote{www.midjourney.com}, which enabled us to retrieve images with a high number of views and likes. 
To ensure the diversity of image content, we adopted the strategy from \cite{haloquest}, combining 17 topic words and 15 attribute words to form query words for retrieving quality images, as shown in Figure \ref{fig:image-topic-words}. This approach results in a significantly larger and more diverse set of topic words for image content compared to previous image-text datasets, which are primarily sourced from MS-COCO \cite{mscoco} or the Flickr platform\footnote{www.flickr.com}.

\section{Compututational Resources}
To run the experiments, we utilized a cluster of A100 GPUs, A40 GPUs, and V100 GPUs. The largest and most resource intensive model we tested, LLaVA-NeXT QWEN-110B, required 4 A100 GPUs for 2 days for the MCOT task while the smallest model we tested, ALBEF-210M, required 1 V100 GPU for 1 hour for the cross-modal retrieval task. On average, depending on the task, all other models were run on 1 V100 GPU for 0-1 hour, or 1-2 A40 GPUs for 2-6 hours, or 1 A100 GPU for 1-3 hours.

\section{Comparison of JourneyBench vs.~JourneyDB and WHOOPS}

There have been limited efforts \cite{whoops, journeydb} to leverage generated images in VLU evaluation. These attempts have not fully exploited the controllability, convenience, and strengths of prompt-based generated images \cite{rombach2022highresolution, midjourney} to address more challenging issues such as MCOT, fine-grained cross-modal retrieval \cite{wang2022paired, tokenflow}, and multi-image visual reasoning \cite{multivqg, mantis, mementos}. Additionally, \cite{whoops} is limited to 500 handcrafted generated images, which not only are vulnerable to human biases in the image creation process but are much constrained in scale. On the contrary, JourneyBench has $13,631$ unique image-text samples across five tasks, which consist of $12,405$ unique images and $13,664$ unique text. Furthermore, \cite{journeydb} are solely annotated by a single model, GPT-3.5, and does not involve any human verification or direct annotation. Thus, it can be vulnerable to model biases and low-quality data. Differently, JourneyBench involves both human-machine-in-the-loop processes to ensure the quality and diversity of our data. Together, our annotators spent more than $2,200$ hours annotating JourneyBench.

\section{Qualitative examples}
\begin{figure}
    \centering
\includegraphics[width=\textwidth, keepaspectratio]{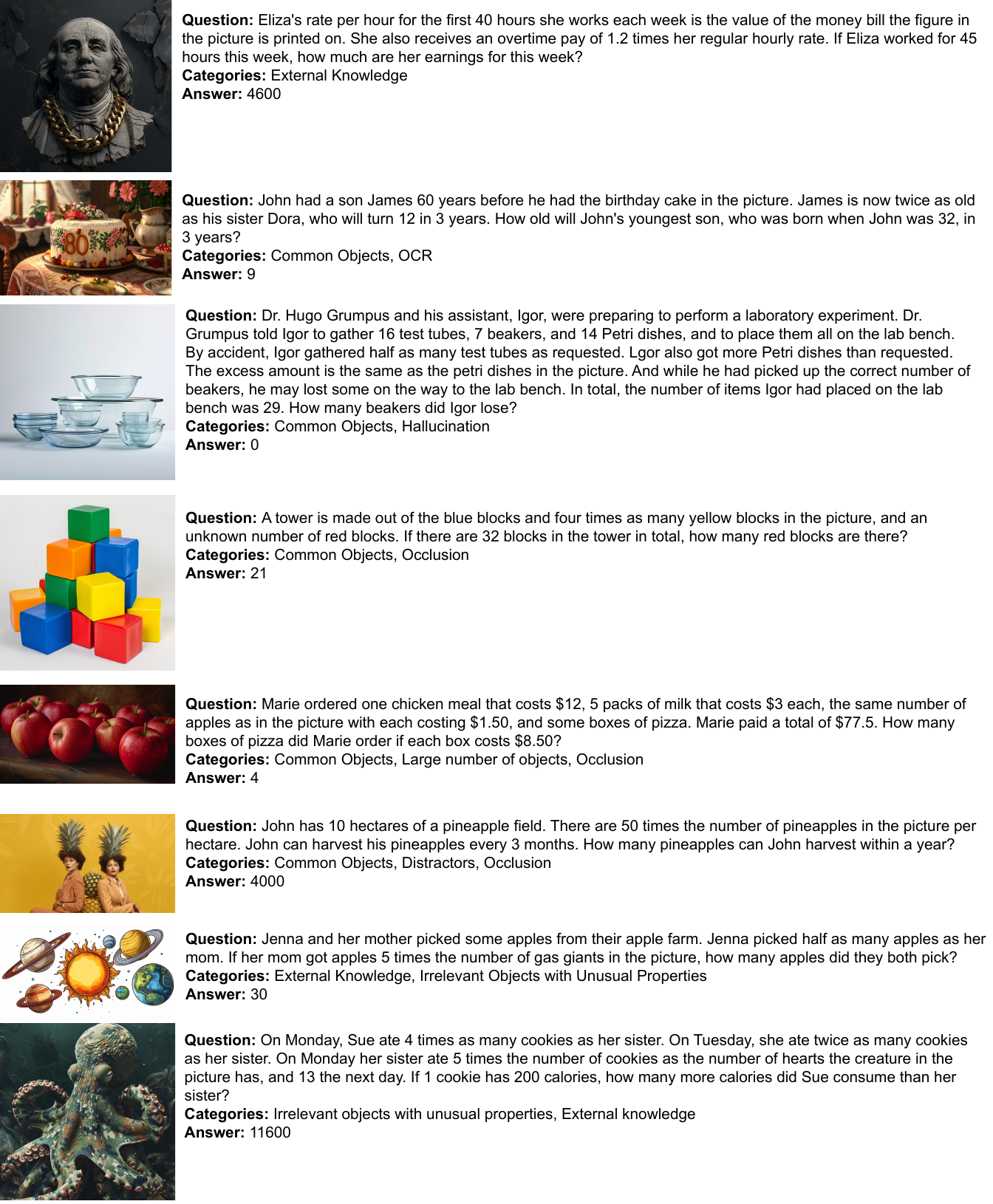}
    \caption{Qualitative examples of MCOT with categories.}
  \label{fig:mcot_example}
\end{figure}
\begin{figure}
    \centering
\includegraphics[width=\textwidth, keepaspectratio]{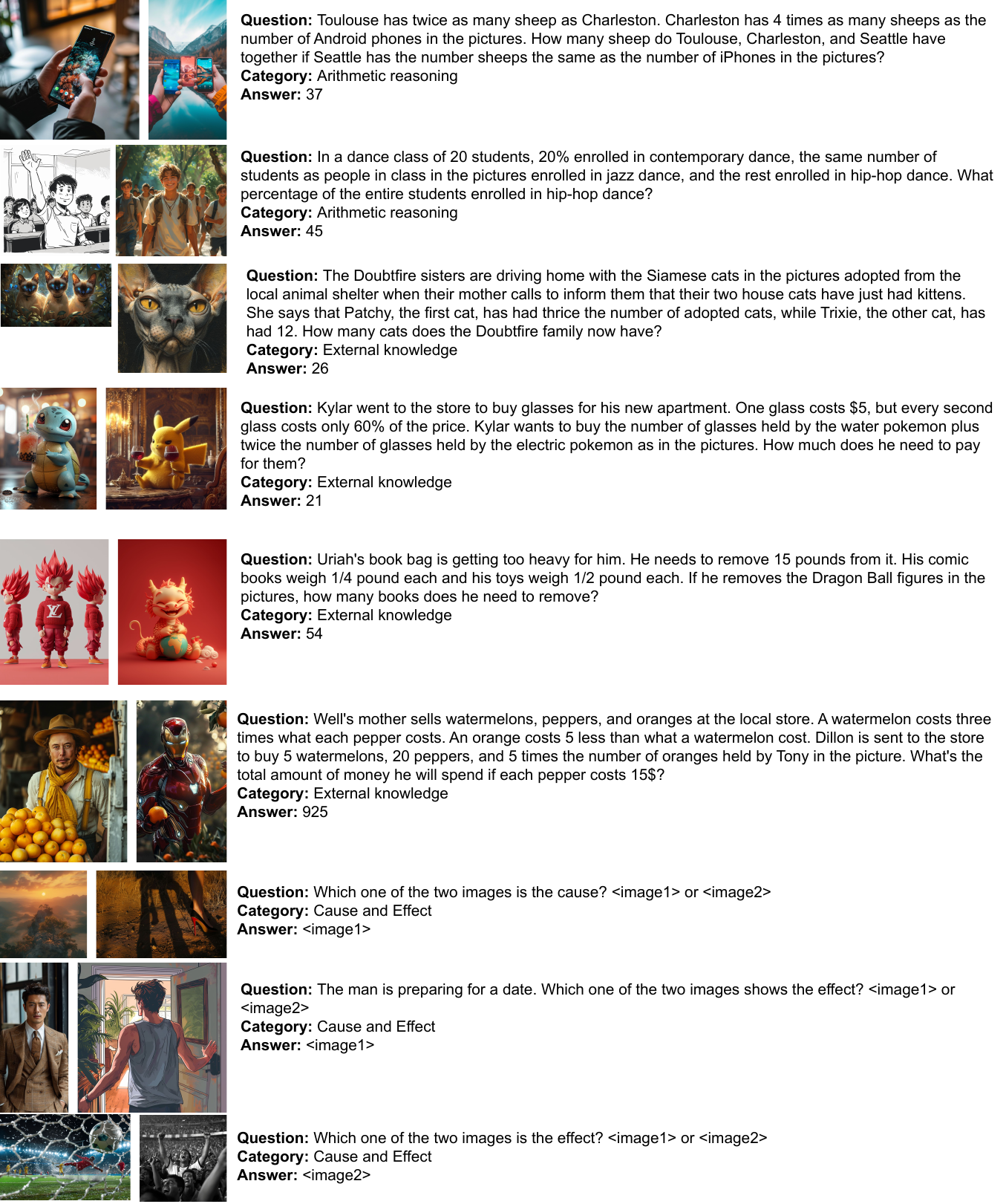}
    \caption{Qualitative examples of Multi-image VQA with categories.}
  \label{fig:multiimage_example}
\end{figure}
\begin{figure}
    \centering
\includegraphics[width=\textwidth, keepaspectratio]{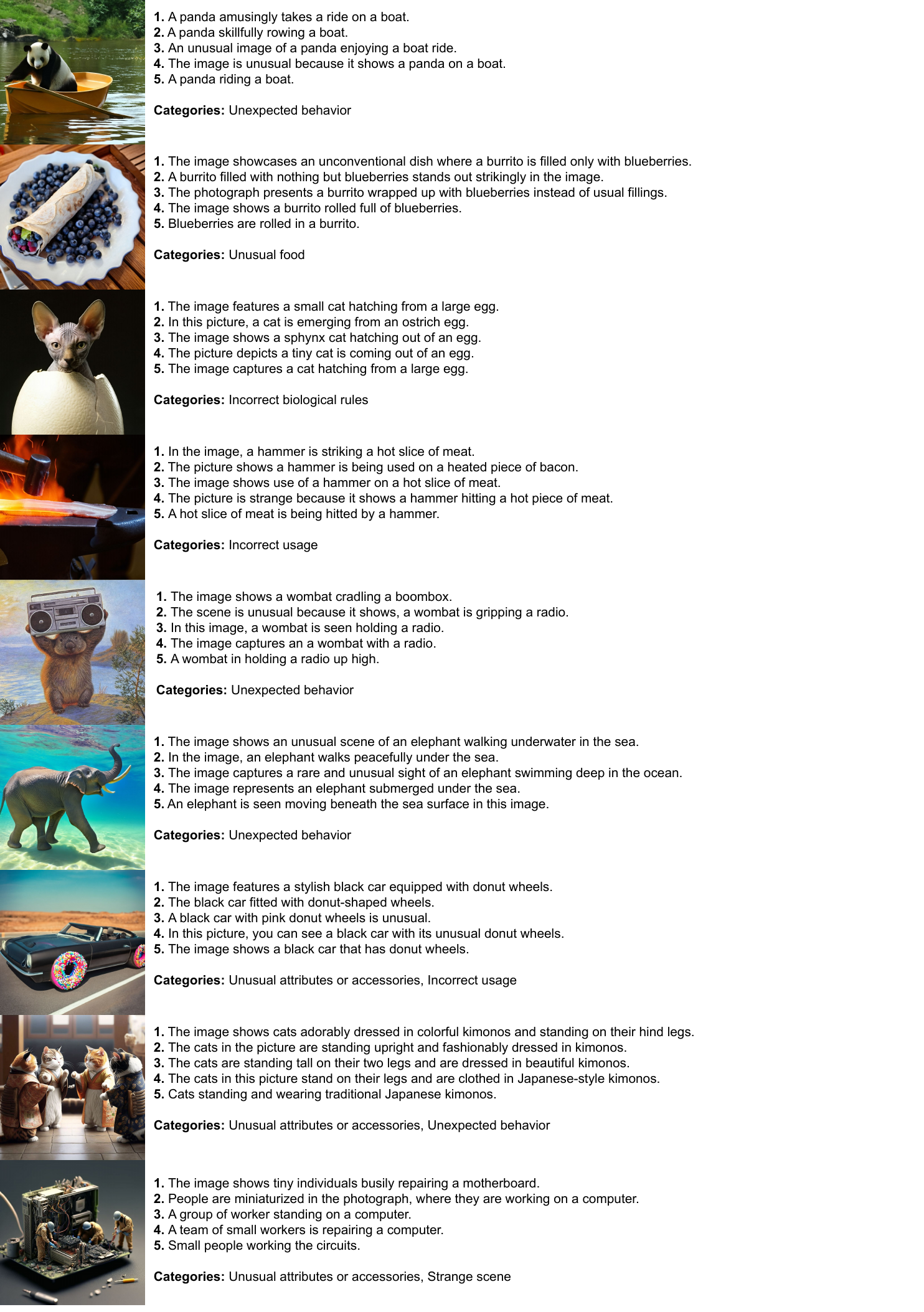}
    \caption{Qualitative examples of Imaginary Image Captioning.}
  \label{fig:ucg_example}
\end{figure}
\begin{figure}
    \centering
\includegraphics[width=\textwidth, keepaspectratio]{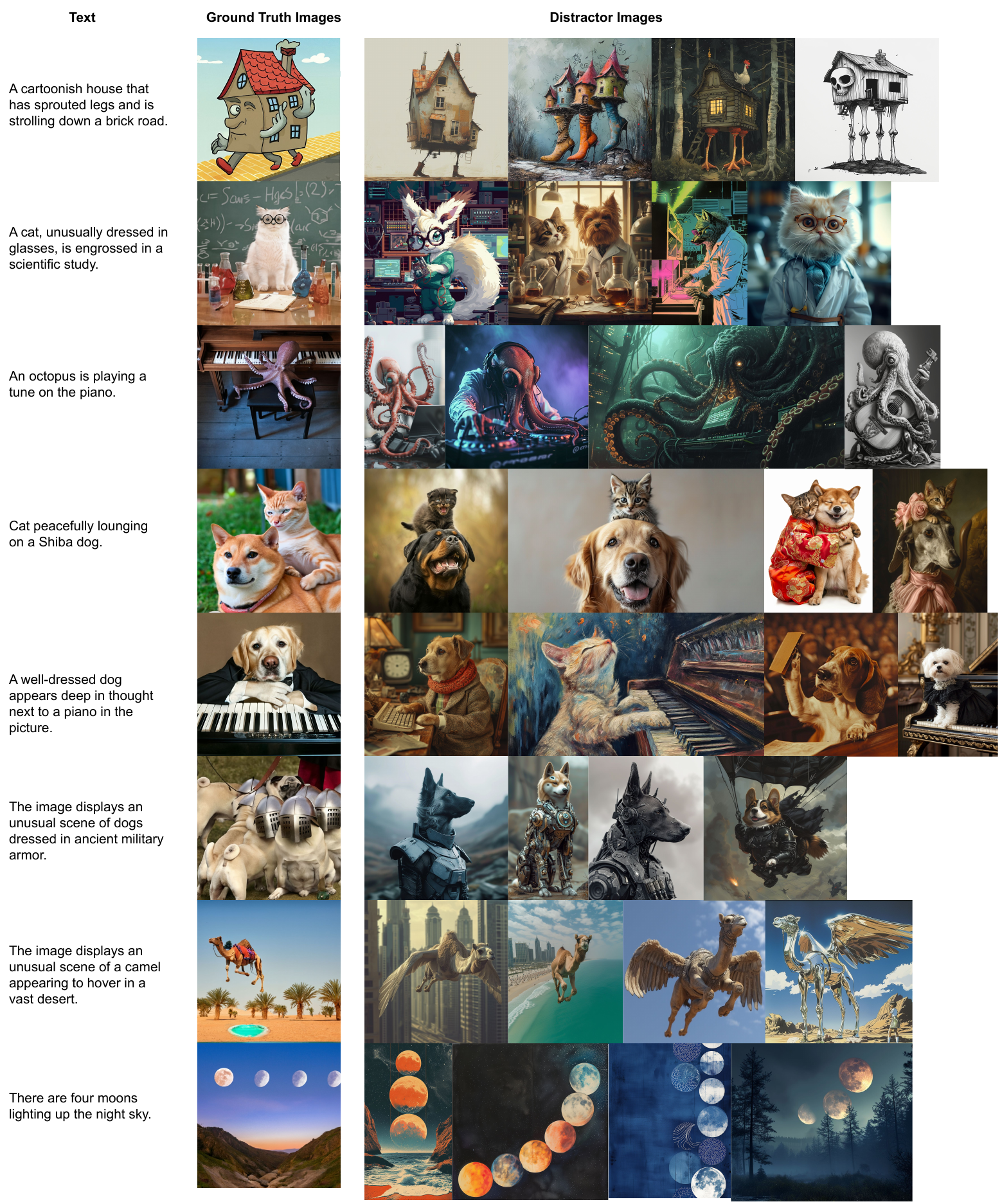}
    \caption{Qualitative examples of text-to-image retrieval with distractors.}
  \label{fig:retrieval_t2i_example}
\end{figure}
\begin{figure}
    \centering
\includegraphics[width=\textwidth, keepaspectratio]{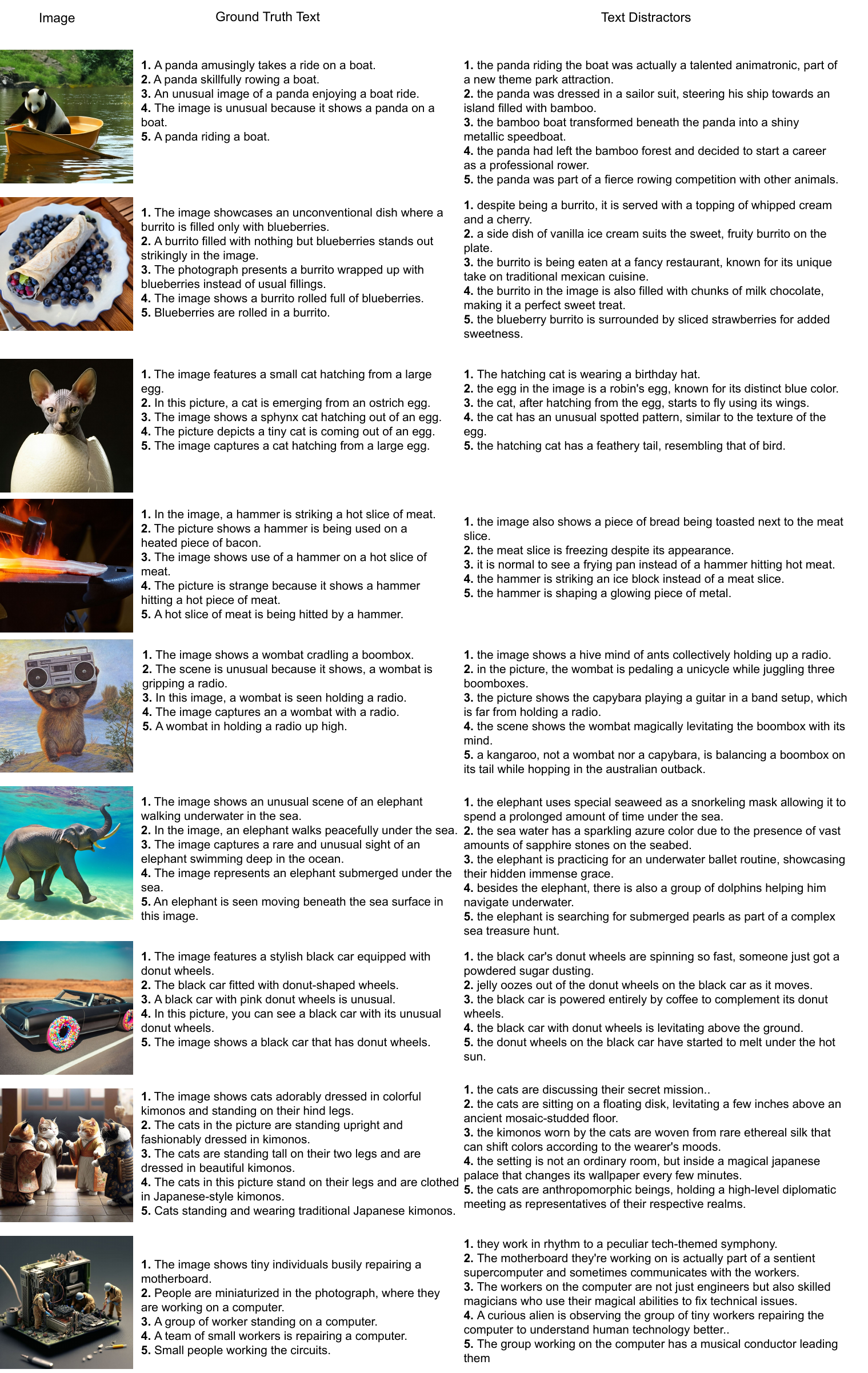}
    \caption{Qualitative examples of image-to-text retrieval with distractors.}
  \label{fig:retrieval_i2t_example}
\end{figure}
\begin{figure}
    \centering
\includegraphics[width=\textwidth, keepaspectratio]{figures_appendix/qualitative_examples/haloquest_examples.pdf}
    \caption{Qualitative examples of HaloQuest.}
  \label{fig:haloquest_example}
\end{figure}
Please refer to Figures \ref{fig:mcot_example}, \ref{fig:multiimage_example}, \ref{fig:ucg_example}, \ref{fig:retrieval_t2i_example}, \ref{fig:retrieval_i2t_example}, and \ref{fig:haloquest_example}

\clearpage
\section{Potential Societal Impacts}
\textit{\textbf{Potential Positive Impacts.}} 
The development and deployment of advanced vision-language benchmarks like JourneyBench have several potential positive societal impacts. 
Firstly, because JourneyBench is a one-stop vision-language benchmark with fine-grained annotations, it makes comparing the performance of different state-of-the-art AI systems easier. For example, JourneyBench exposes that GPT4o has a stronger tendency to hallucinate than GPT4V. Researchers can use JourneyBench to diagnose where models excel and where they struggle to better target their research efforts. JourneyBench thus has the potential to significantly improve the accuracy and reliability of AI systems used in various applications with larger societal benefits, such as medical imaging, autonomous vehicles, and assistive technologies for people with disabilities. 
JourneyBench will allow fairer and broader comparison of AI models by providing a standardized benchmark where models can be compared and improved across a number of axes which have applications in critical downstream applications.
Enhanced accuracy in these domains can lead to better diagnostics, safer transportation, and more effective assistance, thus improving overall quality of life. 
We expect models that will be compared on JourneyBench to be deployed in many sectors, such as intelligent tutoring and question answering.
Further, because the datasets provided by JourneyBench are highly diverse and feature AI generated content, we expect JourneyBench to play an important role in benchmarking performance of AI systems on generated data, which we expect will continue to grow across social media and the internet, as well as benchmarking performance on unusual situations.
Due to its rich diversity of atypical situations and content, JourneyBench will help in creating more robust and less biased AI models which is critical for deployment of AI systems in real world applications. 

\textit{\textbf{Potential Negative Impacts.}} 
On the other hand, there are potential negative societal impacts associated with the use and development of JourneyBench. One major concern is the exacerbation of existing biases within AI systems. 
JourneyBench was harvested from data generated from human prompts on MidJourney with models trained on images harvested from the web.
If the data used to train these models was not carefully curated to avoid reinforcing stereotypes or excluding certain groups, the resulting generated images can perpetuate or even amplify societal inequalities by reflecting those biases within the data.
While JourneyBench was harvested by humans who inspected samples from MidJourney, it is possible that some of these inequities exist within the data, despite being manually chosen (e.g.~overrepresentation of certain racial groups). This may lead to biased analysis of models, such as by overestimating their performance on images containing minorities.
More broadly, JourneyBench will help facilitate the improvement of advanced AI systems which could lead to increased surveillance and erosion of privacy, as more sophisticated AI could be employed in ways that monitor and analyze individuals' behavior without their consent (e.g.~automatially analyzing behaviors, predicting next steps, etc.).
There is also the risk of job displacement in industries where these advanced AI systems are implemented, leading to economic and social challenges for affected workers. 
For example, JourneyBench reveals that many models continue to struggle on multi-image chain of thought reasoning. As these capabilities improve, workers whose roles involve such analysis are at risk of replacement.
To address these issues, we intend to address potential negative impacts through transparency, explicit ethical consideration statements, and policies that ensure AI development aligns with societal values and needs. For example, we will make clear that analysis on JourneyBench may reflect underlying biases.

\section{Limitations}
One primary limitation of JourneyBench is the inherent difficulty in curating truly unbiased and representative imaginary images.
While JourneyBench aims to test models in unusual and imaginative scenarios, the selection of these scenarios might still reflect certain biases or gaps. 
For instance, the types of imaginary images and tasks chosen might not cover all possible edge cases or cultural contexts, potentially limiting the generalizability of the benchmark’s findings.
Additionally, the reliance on generated images, although mitigating copyright issues and enabling diverse content, may introduce artifacts or inconsistencies that are not present in real-world images, potentially skewing the evaluation results.
Because all generated images were harvested from the Midjourney website, generated images may contain biases or artifacts present in the AI models available at this time.
For example, many image generators rely on conditioning from CLIP. If certain visual content is not well captured by CLIP's conditioning, it may not appear in the generated output. 
Further, as generative models advance in the coming years, new classes of models and conditioning may emerge. Those models may contain a different set of artifacts or biases than present in JourneyBench, so performance on JourneyBench may not necessarily translate to those. In particular, some models we evaluate rely on CLIP's conditioning. If CLIP is also used in image generation, this may introduce a bias towards models relying on these encoders. 

Another limitation of JourneyBench is that the tasks within it are designed to be extremely challenging and require complex, fine-grained visual reasoning. This focus on fine-grained details and unusual scenarios may not fully capture the broad utility of these models in more conventional applications, potentially underrepresenting their strengths in real-world tasks.
Other limitations include the focus on English-language understanding (in all captions and question answering tasks), as opposed to other languages. This may further bias JourneyBench towards certain types of content found in English-speaking countries. 
Lastly, JourneyBench does not include any generated video understanding tasks. Prompt-based generated videos can be expected to proliferate in the coming years, with impressive results showcased by OpenAI's SORA. JourneyBench currently focuses on image understanding (including multi-image understanding), but does not currently address temporal understanding in generated videos.

\section{Personally Identifiable Information and Offensive Content}

The JourneyBench dataset is constructed with a strict focus on ethical standards and user safety. It does not contain any personally identifiable information (PII) or sensitive data related to individuals. All images in the dataset are generated and publicly posted for sharing through the Midjourney platform under the \href{https://docs.midjourney.com/docs/terms-of-service}{community rules}, ensuring that no PII is included. Additionally, the dataset has been curated to exclude any content that might be considered offensive, insulting, threatening, or anxiety-inducing. The images underwent a multi-layered filtering process, initially by the Midjourney platform and subsequently through multiple rounds of human annotation, to ensure appropriateness and non-distressful content. This rigorous curation process guarantees that the JourneyBench dataset is suitable for a broad audience and aligns with ethical guidelines for public research and academic use. Therefore, individual consent for data collection is not applicable. The annotations were created by human annotators specifically for research purposes, ensuring that all data within JourneyBench is ethically sourced and suitable for academic and non-commercial research.

\subsection{Digital Object Identifier}
We have requested a DOI for JourneyBench on \url{https://registry.identifiers.org/} and await their approval.

\subsection{HaloQuest Data}

JourneyBench includes a task, VQA with hallucination triggers, which is derived from a previous work titled "HaloQuest: A Visual Hallucination Dataset for Advancing Multimodal Reasoning." HaloQuest is currently under review and planned for release soon. The authors of this work are responsible for both the HaloQuest and JourneyBench data. There are no ethical issues in HaloQuest beyond those already addressed in JourneyBench.

\section{Future Maintenance Plan}

The JourneyBench dataset will undergo regular updates and maintenance to ensure its continued relevance and accuracy in evaluating multimodal models. The research team at Columbia University, UCLA, and Virginia Tech will be responsible for these updates, which will include correcting labeling errors, adding new instances, and removing outdated or erroneous data. Updates will be communicated to users through the official GitHub repository at \url{https://github.com/JourneyBench/JourneyBench}, the project website at \url{https://journeybench.github.io/}, and a mailing list for subscribed users. The team aims to review and update the dataset at least quarterly or more frequently as needed based on feedback and the identification of new challenges in the field. The maintenance would continue for at least five years after the paper's acceptance. Additionally, a leaderboard will be developed to track and document future works and their model performance using the JourneyBench dataset, fostering a collaborative environment for ongoing research and improvement. 

We plan to share the dataset on Hugging Face and host a workshop focusing on a competition via JourneyBench at the upcoming CVPR conference. These initiatives will broaden access to the dataset and encourage active participation and collaboration within the research community.

\section{Terms of Usage for JourneyBench Dataset}

\subsection{Ownership and Responsibility}
The JourneyBench dataset contains images obtained from the Internet, including those generated by Midjourney, which are not the property of Columbia University, UCLA, or Virginia Tech. These institutions are not responsible for the content or meaning of these images.

The authors state that to the best of their knowledge, information, and belief they have obtained all content in JourneyBench from sources such as Midjourney which allow for the intended use and redistribution in JourneyBench. The authors assume full responsibility for violation of any rights from content in JourneyBench and will immediately move to rectify any such violation should such violation be brought to the authors' attention. 
All data was harvested consistent with the Terms of Use of Midjourney and other platforms used by the authors to create and assemble JourneyBench. 

\textbf{\textit{Fair use notice.}} The authors acknowledge that in the United States, copyright of generative content remains an issue in flux. Should any generated content within JourneyBench ever be held to fall under copyright under current US law, JourneyBench can still be distributed under fair use. 
Specifically, we make JourneyBench available in an effort to advance understanding of technological, scientific, and cultural issues. We believe this constitutes a `fair use' of any such copyrighted material as provided for in Section 107 of the US Copyright Law. In accordance with Title 17 U.S.C. Section 107, the material in JourneyBench is distributed without profit to those who have expressed a prior interest in receiving the included information for non-commercial research and educational purposes. For more information on fair use please \href{https://www.law.cornell.edu/uscode/text/17/107}{click here}. If you wish to use copyrighted material in JourneyBench for purposes of your own that go beyond non-commercial research and academic purposes, you must obtain permission directly from the copyright owner should one exist.

\subsection{Non-commercial Research}
The JourneyBench dataset is \textbf{ONLY} available for non-commercial research purposes. Any use of the dataset for commercial purposes is strictly prohibited.

\subsection{Competitive Research}
You may not use the JourneyBench dataset for competitive research against Midjourney or any other image generation platforms.

\subsection{Restrictions on Usage}
\begin{itemize}
    \item You agree not to reproduce, duplicate, copy, sell, trade, resell, or exploit any portion of the images or derived data for commercial purposes.
    \item You agree not to further copy, publish, or distribute any portion of the JourneyBench dataset.
    \item Except for internal use at a single site within the same organization, making copies of the dataset is prohibited.
\end{itemize}

\subsection{Interpretation and Revision}
The research team at Columbia University, UCLA, and Virginia Tech reserves the right to interpret and revise these terms.

\subsection{Removal of Product}
If you do not wish to have your product included in the JourneyBench dataset, please contact us at \href{mailto:journeybench.contact@gmail.com}{journeybench.contact@gmail.com} to have it removed.

By using the JourneyBench dataset, you agree to comply with these terms of usage. Any violation of these terms may result in the termination of your access to the dataset and could lead to legal action.

\subsection{Licensing}
The JourneyBench dataset is distributed under a custom license that includes the following terms based on the Creative Commons Attribution-NonCommercial-NoDerivatives 4.0 International (CC BY-NC-ND 4.0) license, with additional restrictions:
\begin{itemize}
    \item \textbf{Attribution}: You must give appropriate credit, provide a link to the license, and indicate if changes were made. You may do so in any reasonable manner, but not in any way that suggests the licensor endorses you or your use.
    \item \textbf{NonCommercial}: You may not use the material for commercial purposes.
    \item \textbf{NoDerivatives}: If you remix, transform, or build upon the material, you may not distribute the modified material.
    \item \textbf{Additional Restrictions}: The dataset may not be used for competitive research against Midjourney or any other image generation platforms. You also agree not to further copy, publish, or distribute any portion of the dataset beyond internal use at a single site within the same organization.
\end{itemize}

For more details, visit \url{https://creativecommons.org/licenses/by-nc-nd/4.0/}.

By incorporating these terms, the JourneyBench dataset can be distributed in a manner that respects the privacy and usage policies of the original sources, while also ensuring it is used appropriately within the research community.

\end{document}